%% file: main.tex
\documentclass[11pt, a4paper, copyright, goog]{google}

\usepackage[authoryear, sort&compress, round]{natbib}
\usepackage{hyperref}
\usepackage{url}
\usepackage{multirow}
\usepackage{booktabs}
\usepackage{graphicx}  

\usepackage{microtype}
\usepackage{graphicx}
\usepackage{booktabs} 

\usepackage{hyperref}
\usepackage{url}            
\usepackage{booktabs}       
\usepackage{amsfonts}       
\usepackage{nicefrac}       
\usepackage{microtype}      
\usepackage{xcolor}         
\usepackage{wrapfig,lipsum,booktabs,graphicx}
\usepackage{amsmath}
\usepackage{amsthm}
\usepackage{subfig}
\usepackage{wrapfig,lipsum,booktabs,graphicx}
\usepackage{tikz}
\usepackage{framed}
\usepackage{graphicx}
\usepackage{array}
\usepackage{hhline}
\usepackage{subcaption}
\usepackage{color}
\usepackage{multirow}
\usepackage[noend]{algorithmic}
\usepackage{algorithm}

\usepackage{amsmath}
\usepackage{amssymb}
\usepackage{mathtools}
\usepackage{amsthm}

\usepackage[capitalize,noabbrev]{cleveref}

\theoremstyle{plain}

\theoremstyle{definition}

\theoremstyle{remark}

\usepackage[textsize=tiny]{todonotes}

\usepackage[utf8]{inputenc} 
\usepackage[T1]{fontenc}    
\usepackage{hyperref}       
\usepackage{url}            
\usepackage{booktabs}       
\usepackage{amsfonts}       
\usepackage{nicefrac}       
\usepackage{microtype}      
\usepackage{xcolor}         


\uselogo{} 

\title{You Only Fine-tune Once: Many-Shot In-Context Fine-Tuning for Large Language Models}

\correspondingauthor{wenchong@google.com, liqianp@google.com}

\reportnumber{0001} 

\author[*,1]{Wenchong He}
\author[*,1]{Liqian Peng}
\author[2]{Zhe Jiang}
\author[1]{Alec Go}

\affil[*]{Equal contributions}
\affil[1]{\thepa{}{}}
\affil[2]{University of Florida}

\begin{abstract}
Large language models (LLMs) possess a remarkable ability to perform in-context learning (ICL), which enables them to handle multiple downstream tasks simultaneously without requiring task-specific fine-tuning. Recent studies have shown that even moderately sized LLMs, such as Mistral 7B, Gemma 7B and Llama-3 8B, can achieve ICL through few-shot in-context fine-tuning of all tasks at once. However, this approach still lags behind dedicated fine-tuning, where a separate model is trained for each individual task.
 In this paper, we propose a novel approach, Many-Shot In-Context Fine-tuning (ManyICFT), which significantly narrows this performance gap by extending the principles of ICL to a many-shot setting. To unlock the full potential of ManyICFT and address the inherent inefficiency of processing long sequences with numerous in-context examples, we propose a novel training objective.  Instead of solely predicting the final answer, our approach treats every example within the context as a supervised training target. This effectively shifts the role of many-shot examples from prompts to targets for autoregressive learning.  Through extensive experiments on diverse downstream tasks, including classification, summarization, question answering, and natural language inference, we demonstrate that ManyICFT substantially outperforms zero/few-shot fine-tuning and approaches the performance of dedicated fine-tuning.  Furthermore, ManyICFT significantly mitigates catastrophic forgetting issues observed in zero/few-shot fine-tuning. The code will be made publicly available on publication. 
\end{abstract}

\begin{document}

\maketitle

\section{Introduction}
\label{intro}
Fine-tuning large language models (LLMs) for specific downstream applications is a time-consuming  and resource-intensive process, hindering rapid scalability \citep{NEURIPS2020_1457c0d6,xu2023parameter}. 
The constant emergence of new base models, such as daily updates in Hugging Face, further exacerbates this challenge \citep{touvron2023llama}.  To address this, we explore an \textit{all-at-once fine-tuning } approach, where a single base model is jointly fine-tuned on multiple tasks simultaneously. This results in a versatile, fine-tuned model capable of performing well on a wide variety of downstream tasks, including unseen ones, without requiring any further task-specific fine-tuning.

\begin{figure}
    \centering
    \includegraphics[width=1.8in]{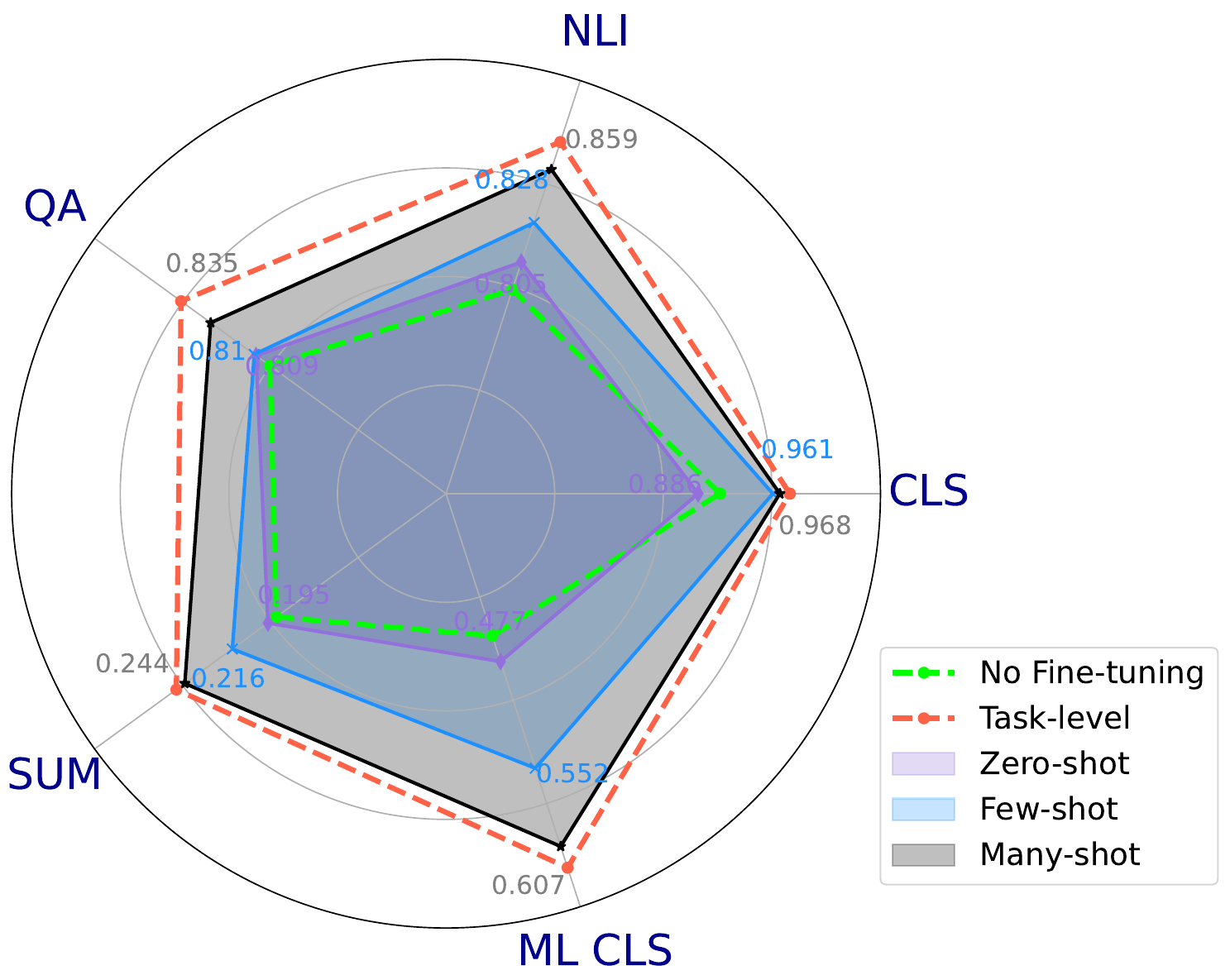}
\caption{ {\bf Comparison of different fine-tuning strategies.} Many-shot fine-tuning for ICL (solid \textbf{black} line) achieves performance comparable to task-level LoRA fine-tuning (dotted \textcolor{red}{red} line, five different models). The tasks include Classification (CLS), Multilingual Summarization (SUM), Question Answering (QA), Natural Language Inference (NLI). Multi-label classification (ML CLS)}
    \label{fig:teaser}
\end{figure}

In-context learning (ICL) offers an elegant approach to adapting LLMs for downstream tasks by simply providing a few examples, eliminating the need for explicit fine-tuning. However, ICL, particularly with moderately sized LLMs (2B to 13B parameters), often lags behind dedicated fine-tuning, especially for complex  domain-specific tasks. While recent work on few-shot in-context fine-tuning \citep{Min2022metaicl}  attempted to bridge this gap, a clear performance discrepancy remains.

Inspired by the success of many-shot in-context learning (without fine-tuning) in LLMs with extensive context \citep{team2023gemini, yang2024qwen2}, we propose Many-shot In-Context Fine-Tuning (ManyICFT), which extends in-context fine-tuning from a \textit{few-shot}  to a \textit{many-shot} setting for moderately sized models (around 7B parameters).  ManyICFT also  addresses   the  inherent   inefficiency of processing long sequences with numerous examples by introducing a novel training \begin{wrapfigure}{r}{7.7cm} 
    \centering
    \includegraphics[width=2.9in]{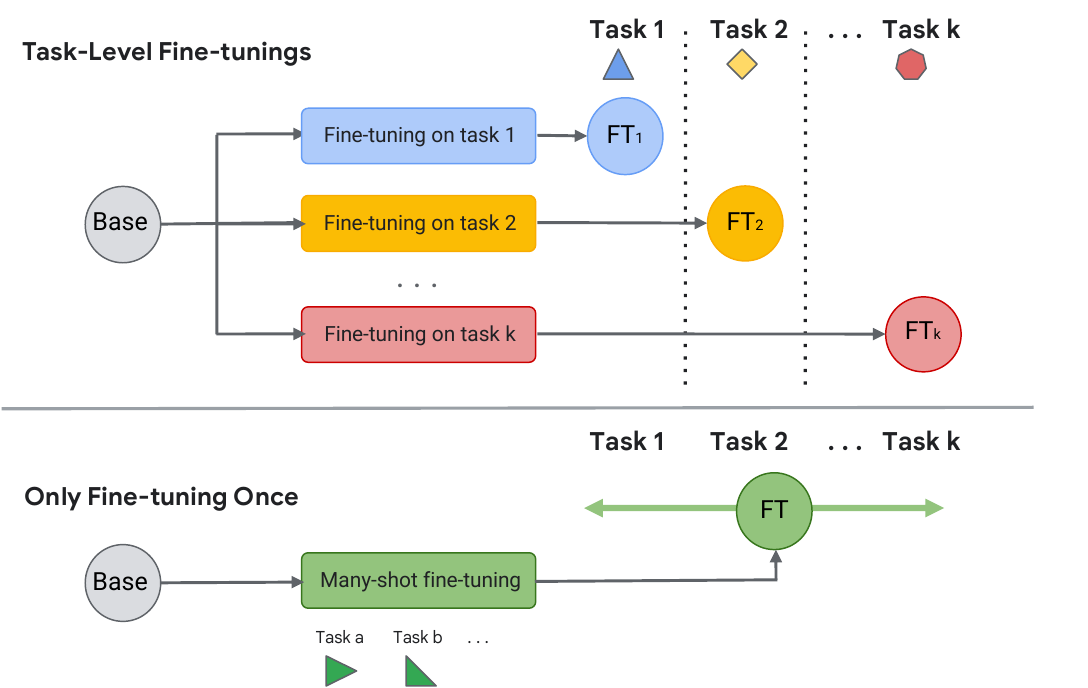}
    \caption{{\bf Comparison between ManyICFT and task-level fine-tuning workflows.} Task-level fine-tuning requires maintaining a separate model for each downstream task, whereas ManyICFT adapts effectively to unseen datasets using many-shot prompting within a single model (``FT" is the fine-tuned model, ``Base" is the base LLM model).  }
    \label{fig:workflow}
\end{wrapfigure} objective. Instead of solely predicting the final answer, ManyICFT treats every example within the context as a prediction target, effectively shifting the role of many-shot examples from prompts to targets for autoregressive learning. This technique significantly mitigates the gap towards the dedicated fine-tuned models (fine-tuned on in-domain dataset), thus offering an alternative development workflow. As depicted in Figure \ref{fig:workflow}, ManyICFT enables LLMs to be adapted to downstream tasks through many-shot prompting, including those unseen datasets. This eliminates the need to maintain separate fine-tuned models for each task (See Table~\ref{tab:compcosts}). Consequently, ManyICFT has significant potential to streamline and simplify the downstream development process.

Our experiments on held-out datasets across five downstream tasks demonstrate that ManyICFT outperforms both few-shot and zero-shot fine-tuned models and achieves performance levels comparable to task-specific fine-tuning. The results are shown in Figure \ref{fig:teaser}. Compared to state-of-the-art few-shot fine-tuning, ManyICFT improves in-context performance by $1.3\%$ in classification, $3.1\%$ in natural language inference, $2.5\%$ in question answering,  $2.0\%$ on summarization and $4.2\%$ in multi-label classification. These enhanced outcomes demonstrate that ManyICFT achieves performance comparable to dedicated task-specific fine-tuned models.

In conclusion, our \textbf{main contributions} are:
  \begin{itemize}
\item \textbf{Extend in-context fine-tuning from few-shot to many-shot:} We present a new training objective \textit{mask all targets}, which significantly improves the training efficiency for many-shot fine-tuning. This facilitates the extension of in-context fine-tuning from few-shot to many-shot settings. 
\item \textbf{Enhanced ICL performance:} Our evaluations demonstrate that ManyICFT enhances performance in multiple downstream tasks, outperforming the base model, zero/few-shot fine-tuning, and greatly reduces the gap between ICL and task-level dedicated fine-tuning. 
\item \textbf{Mitigate catastrophic forgetting:} Evaluations of  long-context capability and ablation study on out-of-domain generalization demonstrate that ManyICFT substantially mitigates catastrophic forgetting problems observed in zero-shot and few-shot fine-tuning. 



\end{itemize}

\section{Related Work}
 \subsection{In-Context Learning (ICL)}

ICL enables language models to quickly adapt to new situations based on the provided context, making them versatile and powerful tools for a wide range of natural language processing tasks. ICL offers several advantages over fine-tuning. First, because demonstrations are written in natural language, ICL provides an interpretable interface for interacting with language models \citep{brown2020language}. This approach allows for the easy incorporation of human knowledge into LLMs by simply modifying the demonstrations and templates \citep{liu-etal-2022-makes,wei2022finetuned}. Second, ICL mirrors the human decision-making process, which often relies on learning by analogy \citep{winston1980learning}. Third, unlike supervised training, ICL is a training-free learning framework. This not only significantly reduces the computational costs associated with adapting the model to new tasks but also offers flexibility, as it can easily adapt to new tasks without retraining. This flexibility facilitates the implementation of language-model-as-a-service \citep{sun2022black}, making ICL particularly well-suited for large-scale real-world applications.

Recent advancements in long-context LLMs have led to increased research on the impact of ICL with a growing number of demonstration examples. For instance, Gemini 1.5, with a 2 million token context window, has demonstrated strong many-shot ICL capabilities \citep{Agarwal2024many}. However, lighter or medium-sized LLMs (with 2-20 billion parameters) still face challenges with many-shot ICL. For example, \cite{Li2024long} evaluated several 7 billion parameter LLMs with a 32K token context length on classification tasks and found that many-shot ICL  does not consistently improve performance, particularly on more challenging problems. There is a need to improve the in-context learning capability for lighter or medium-sized LLMs. 



\subsection{Meta Learning for Few-Shot ICL}

Since most pretraining data are not specifically tailored for ICL \citep{chen2022improving}, various  fine-tuning strategies have been developed to bridge the gap between pretraining and ICL inference. For instance, \cite{Min2022metaicl} proposed continually fine-tuning LLMs on a diverse range of tasks with multiple demonstration examples, thereby enhancing their ICL capabilities. Additionally, several studies have emphasized the importance of instruction-based methods \citep{mishra2021natural}, \citep{wei2022finetuned}. The FLAN framework \citep{wei2022finetuned} improves LLMs' ability to follow instructions, enhancing both zero-shot and few-shot ICL performance. Furthermore, \cite{chung2022scalinginstructionfinetunedlanguagemodels} and \cite{wang-etal-2022-super} have advocated for scaling instruction tuning with over 1000 task instructions. However, it remains unclear whether these methods can effectively enhance many-shot learning capabilities. For a comprehensive overview of these methods, refer to the survey paper by \cite{dong2024surveyincontextlearning}.


\subsection{Multi-Task Learning}
Recent work has explored multi-task Low-Rank Adaptation (LoRA), to improve the performance of a single LLM on multiple tasks. For example, MoDULA [1] introduces a Mixture-of-Experts (MoE) framework that combines "universal" LoRA modules with domain-specific ones, using a routing network to select experts for a given input. Similarly, MTL-LoRA [2] focuses on optimizing the LoRA structure itself to better disentangle and share representations across tasks. More recently, MoRE and  MiLoRA [3, 4] proposed a prompt-aware routing mechanism where a single expert is selected per prompt to improve multi-task training performance. While these methods are effective for the set of tasks they are trained on, their primary goal is to optimize performance by explicitly encoding multi-task capabilities into the model's architecture or parameters. Their ability to generalize to entirely unseen tasks is limited.

In contrast, ManyICFT's meta-training objective is to teach the LLM how to effectively learn from in-context examples and enhance its generalization capability to unseen domain in training. By fine-tuning the model to leverage a large number of in-context examples, ManyICFT enhances the model's fundamental in-context learning ability. This "fine-tune once" approach results in a single  model that can be adapted to a broad array of downstream tasks, including unseen ones, simply through many-shot prompting at inference time. This eliminates the need for further task-specific tuning.

\section{Many-shot In-context Fine-tuning}
ICL empowers language models to acquire new tasks by leveraging a few examples provided as demonstrations. To enhance many-shot ICL capabilities, we propose fine-tuning the model using explicit many-shot in-context examples. Our hypothesis is that upstream fine-tuning  enables the model to more effectively leverage these in-context examples, and that these capabilities can transfer to downstream, out-of-domain datasets. This extends MetaICL, i.e., employing a multi-task learning scheme across a large collection of meta-training tasks, from few-shot ICL \citep{Min2022metaicl} to many-shot ICL, allowing the model to learn how to condition on a  set of examples, understand the task's semantics, and generate the appropriate output. In the following, we introduce the details for many-shot in-context fine-tuning (ManyICFT).

\subsection{Mask Last Target}

As outlined in previous work \citep{brown2020language}, training examples are concatenated into a single input sample (i.e., a  sequence) for the model, which is particularly well-suited for $n$-shot learning. When $n = 0$, it is referred to as zero-shot learning, which is equivalent to standard supervised fine-tuning (SFT). For $1 \leq n \leq 20$, it is considered few-shot learning; in this paper, we selected $n = 5$ for our experiments following existing work setup \citep{wei2021finetuned, brown2020language}. When $n > 20$, it is termed many-shot learning. For our experiments, the maximum number of examples $n$  approximately varies from $20$ to $1500$ depending on the average token length of the dataset. 
\begin{figure*}[h]
    \centering
    \subfloat[Mask last target]{ \includegraphics[height=1.3in]{ 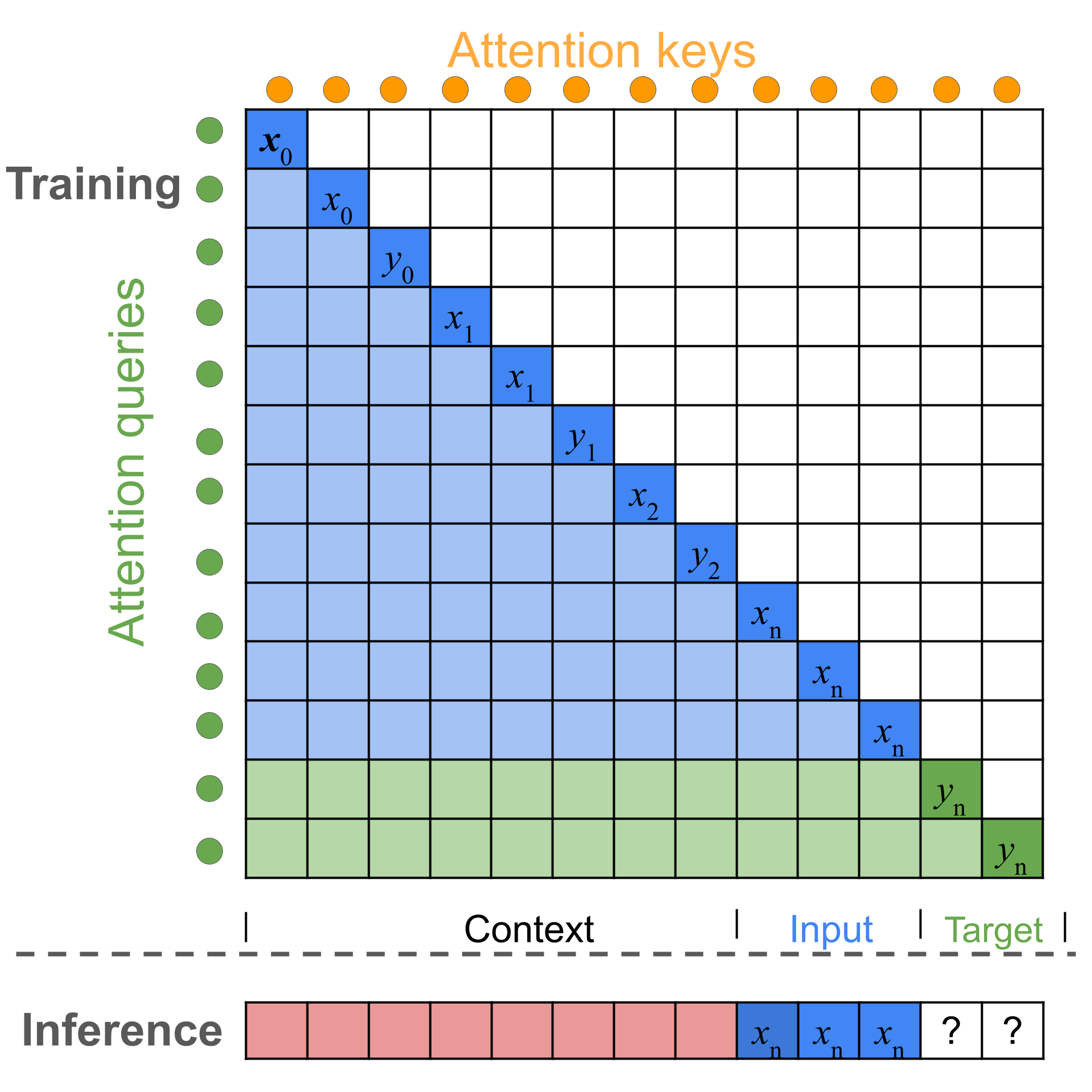}} 
    \subfloat[Mask all targets]{\includegraphics[height=1.3in]{ 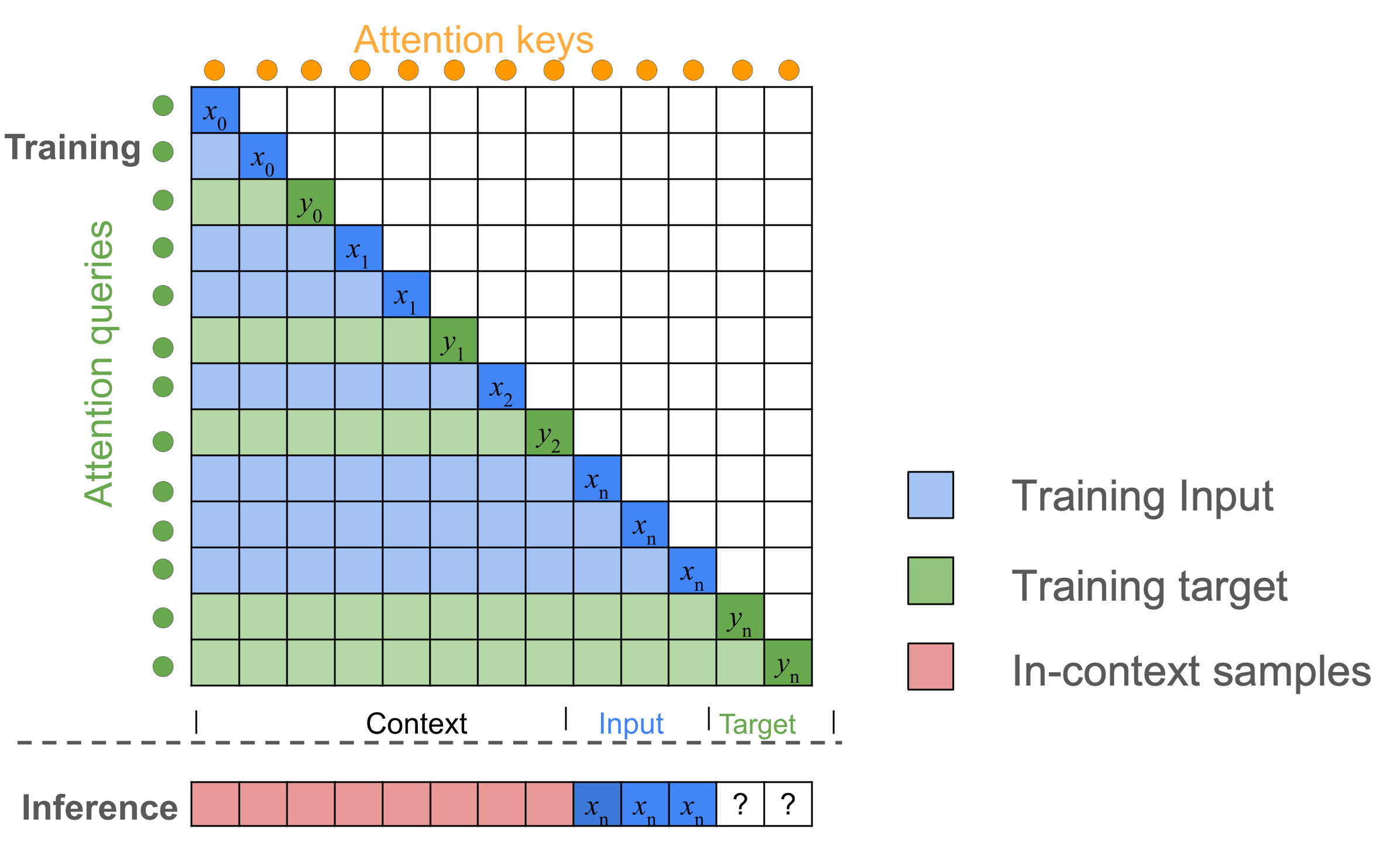}}
    \caption{{\bf Comparison of attention mechanisms  between mask-last-target and mask-all-targets.} The blue  represents the input prompts and green  represents the target outputs. The red square in the inference grid denotes the in-context examples.}
    \label{fig:mask_attention}
\end{figure*}

\textbf{Training:} For many-shot learning, instead of using a fixed value of $n$ for each task, we propose the \textit{maximum in-context length} approach, where $n$ is maximized to fully utilize the available context length during training, thereby minimizing padding. Let $\{x_i, y_i\}_{i=0}^{n-1}$ denote $n$ input-output pairs used as prompts. To enhance the model's in-context learning (ICL) capability, we fix all the examples in the prompt and use $x_n$ as the input, with the corresponding ground truth $y_n$ serving as the label for fine-tuning the LLM. This method allows the model to learn effectively from in-context examples, utilizing the longest possible context.

\textbf{Inference:} The model is evaluated on an unseen target dataset that includes $n$ training examples, with inference conducted using the same data format as in meta-training.  Note that the same many-shot prompt can be reused for the same task. To optimize inference time, the prompt is processed once and then cached, utilizing transformer key value (KV) caching to avoid redundant processing of the long context. Figure \ref{fig:mask_attention} (a) gives an overview of this approach. For simplicity, any additional context or task prompts are omitted in the figure. 

\subsection{Mask All Targets}
While the \textit{mask last target} training strategy with  \textit{maximum in-context length}  can significantly enhance many-shot ICL capabilities (see the experimental section for more details), it has two main limitations: 

 \textit{Limited Few-Shot Optimization}: This strategy is specifically designed for $n$-shot optimization and does not necessarily improve performance for fewer-shot scenarios. In contrast, models such as MetaICL are optimized for few-shot prompts and may perform better in those contexts.

 \textit{Low Training Efficiency}: In zero-shot or few-shot ICL training, each sequence is typically much smaller than the context length. By concatenating multiple short samples into a single long input sequence, we can increase the number of samples per context window and make more efficient  use of the model's context length. However, in many-shot ICL scenarios where each sample already occupies the entire context length, each context window will contain only one sample. Consequently, training can be slower and generally requires significantly more steps to converge to the desired training loss.

\begin{wrapfigure}{r}{7.7cm} 
    \centering
    \includegraphics[width=2.5in]{ 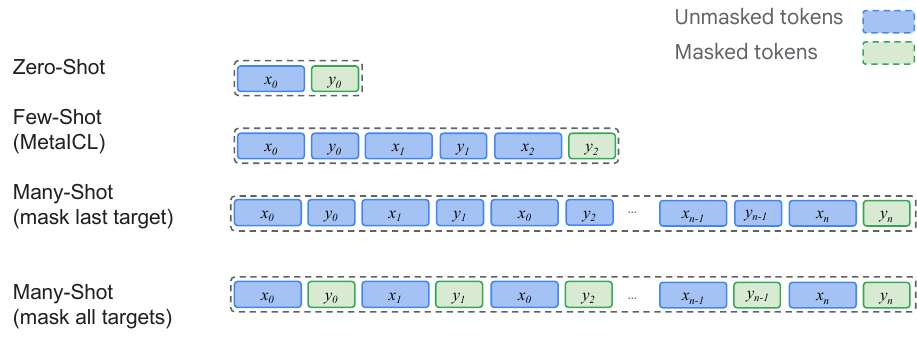}
    \caption{{\bf Illustration of various fine-tuning approaches for ICL models.} The figure shows zero-shot, few-shot, and many-shot fine-tuning methods. For both few-shot ($n=2$ here) and many-shot ICL, we can apply either  \textit{mask last target} or \textit{mask all targets} strategy.}
    \label{fig:demo}
\end{wrapfigure}

To simultaneously enhance the ICL capabilities of few-shot and many-shot, we propose the \textit{mask all targets} training strategy, where all labels are masked and contribute to the loss function (as  shown in Figures \ref{fig:mask_attention} and \ref{fig:demo}). This approach is similar to BERT \citep{Devlin2018BERT}, where masked tokens include   not only the last few positions but also those in the middle of the sequence.  However, unlike BERT, which applies masks to random tokens, our strategy masks only target tokens. When we mask $y_0$, we are training zero-shot inference capability; when we mask $y_k$ for some $0<k\le20$, we are training few-shot ICL capability; and when we mask a larger number of targets $k>20$, we are focusing on many-shot ICL.

\subsection{New Capabilities}

The training strategies presented in this paper focus on optimizing many-shot ICL capabilities. However, we have found that the fine-tuned model, ManyICFT, also demonstrates additional capabilities:

 \textbf{Simplified deployment}: The performance of ManyICFT not only exceeds that of zero and few-shot models, but also achieves comparable performance with task-level LoRA fine-tuning across a range of downstream tasks (as shown in Figure \ref{fig:teaser}). By leveraging many-shot prompts, ManyICFT eliminates the need for additional fine-tuning on downstream tasks, significantly reducing both computational overhead and deployment complexity in industrial settings (as shown in Figure~\ref{fig:workflow} and Table~\ref{tab:numofexamples}). 
 

\textbf{Preserve long context performance}: Most fine-tuning datasets contain samples that are significantly shorter than the available context window, and fine-tuning with customized data can further reduces the long context capability of the base model. Our many-shot fine-tuning approach offers a simple yet effective method to preserve long context capabilities using widely available open-source datasets (see the section 4.2).

\subsection{Computational Analysis}

{\color{black} {\bf Efficient training}: As shown in Table~\ref{tab:training-strategies}, the mask all targets training strategy effectively simulates a mixture of $0$ to $n$ shot learning scenarios within a single batch.  This is achieved by reusing the same token sequence $n$ times, leading to significant reductions in training costs. Let $n$ denote the average number of in-context examples and let $n_w$ represents the maximum number of tokens in the context window. In mask last target strategy, learning across $0$ to $n$ shots independently requires  $n$ separate training instances, resulting in a token complexity of $O(n \cdot n_w)$.  In contrast, the mask-all targets consolidates all in-context examples into a single training instance by masking all target positions simultaneously. This reduces the  token complexity to $O(n_w)$, avoiding redundant computation across similar sequences. Thus, this enables efficient learning from zero-shot to many-shot settings.

{\color{black} {\bf Inference with KV cache}: Let $N_1$ denote the length of the in-context prompt and $N_2$ the length of the query and the generated output, giving a total sequence length $N = N_1 + N_2$. For standard inference without caching, it scales quadratically $O(N^2)$.  With  KV caching, the computation reduces to $O(N_1 \cdot N_2)$. In typical in-context learning settings, $N_1 \gg N_2$, KV caching provides significant efficiency during inference.

The practical computational costs are detailed in Appendix Sections B.1 and B.2. A summary of the key findings is presented in Table~\ref{tab:compcosts}. Compared to conventional task-specific fine-tuning, the ManyICFT meta-training approach achieves a 14 times reduction in training tokens due to the benefits of many-shot meta-learning, and a 13 times decrease in total development time. In addition, the use of the mask-all training strategy provides further computational efficiency. While the inference complexity of ManyICFT is higher, it remains practically efficient, and achieves up to 100 times faster inference when leveraging KV caching.


\begin{table*}[h!]\scriptsize
\centering
\caption{{\bf Comparison of mask-all and mask-last target training strategy.} $n$ is the maximum number of shots that fit within a single context window, $n_w$ is the length of the maximum context window, and $n_x$ and $n_y$ represent the average lengths of the input and target within each example, respectively (with masked targets indicated in \textcolor{blue}{blue}).}
\begin{tabular}{>{\centering\arraybackslash}m{3cm}>{\raggedright\arraybackslash}m{5.5cm}>{\centering\arraybackslash}m{3cm}>{\centering\arraybackslash}m{1.5cm}}
\hline
Training Strategy & Training Input & All Tokens \\
\hline
Mask All Targets & $x_0,  \textcolor{blue}{y_0}, x_1,  \textcolor{blue}{y_1}, x_2,  \textcolor{blue}{y_2}, \dots, x_n, \textcolor{blue}{y_n}$  & $O(n_w)$ \\
\hline
Mask Last Target & 
\begin{tabular}[c]{@{}c@{}}
$x_0,  \textcolor{blue}{y_0}$ \\
$x_0, y_0, x_1,  \textcolor{blue}{y_1}$ \\
$x_0, y_0, x_1, y_1, x_2,  \textcolor{blue}{y_2}$ \\
$\dots$ \\
$x_0, y_0, x_1, y_1, x_2, y_2, \dots, x_n, \textcolor{blue}{y_n}$
\end{tabular} 
 & $O(n_w \cdot n)$ \\
\hline
\end{tabular}
\label{tab:training-strategies}
\end{table*}


\begin{table*}[h!]\scriptsize
\centering
\caption{\textbf{Comparison of Task-Level Fine-Tuning and ManyICFT Meta-Training.} Overview of training and inference efficiency between multiple task-specific fine-tunings and a single ManyICFT meta-trained model. With ManyICFT meta-training, we \textit{only fine-tune once}, and directly use ICL prompt to adapt to $N$ downstream inference tasks ( $T_{\text{SFT}}$ denotes costs of supervised fine-tuning for one task, $T_{\text{META-SFT}}$ is the costs of ManyICFT meta training, $m$ is the number of training instances for task-level fine-tuning, $n_t$ is the maximum context length for task-level fine-tuning).}
\begin{tabular}{|c|c|c|}
\hline
\textbf{\begin{tabular}[c]{@{}c@{}}Category\end{tabular}} &
\textbf{\begin{tabular}[c]{@{}c@{}} Task-Specific Fine-Tuning \\ (Multiple SFTs)\end{tabular}} &
\textbf{\begin{tabular}[c]{@{}c@{}} ManyICFT Meta-Training \\ (One SFT)\end{tabular}} \\
\hline

\begin{tabular}[c]{@{}c@{}}Handling \\ New Tasks\end{tabular} &
\begin{tabular}[c]{@{}c@{}} Fine-tune new \\ LoRA adapters  per task\end{tabular} &
\begin{tabular}[c]{@{}c@{}}Adapt via ICL  prompting \\ with existing LoRA\end{tabular} \\
\hline

\begin{tabular}[c]{@{}c@{}}No. of Adapters \\ ($N$ new tasks)\end{tabular} &
\begin{tabular}[c]{@{}c@{}}$N$ (Typical 1K)\end{tabular} &
\begin{tabular}[c]{@{}c@{}}1\end{tabular} \\
\hline

\begin{tabular}[c]{@{}c@{}}Training Time \\ (Details in Appendix B.1.1)  \end{tabular} &
\begin{tabular}[c]{@{}c@{}}$N \times T_{\text{SFT}}$ \\ (Typical 1K $\times$ 1h =  1000 hours)\end{tabular} &
\begin{tabular}[c]{@{}c@{}}$T_{\text{META-SFT}} $ \\ (Typical 70 hours)\end{tabular} \\
\hline

\begin{tabular}[c]{@{}c@{}}Training  Tokens   \\ (Details in Appendix B.1.2)  \\ \end{tabular} &
\begin{tabular}[c]{@{}c@{}}  $N\times n_t \times m$ \ (Typical 1K $\times$ 4K $\times$ 8K  = 32B)\end{tabular} &
\begin{tabular}[c]{@{}c@{}}$N_{\text{META}}\times n_w$ (Typical 70K $\times$ 32K  = 2.2B)\end{tabular} \\
\hline

\begin{tabular}[c]{@{}c@{}}Inference  Complexity  \\ (Details in Appendix B.2.1) \end{tabular} &
\begin{tabular}[c]{@{}c@{}}${O}(N_2\times N_2)$\end{tabular} &
\begin{tabular}[c]{@{}c@{}}${O}(N_1 \times N_2)$\end{tabular} \\
\hline

\begin{tabular}[c]{@{}c@{}}Relative Inference  Time  \\ (Details in Appendix B.2.2) \end{tabular} &
\begin{tabular}[c]{@{}c@{}}$0.8\times$\end{tabular} &
\begin{tabular}[c]{@{}c@{}}$1\times$ (with KV cache)\end{tabular} \\
\hline
\end{tabular}
\label{tab:compcosts}
\end{table*}
\section{Experiments}
\label{sec:exp}

This section compares our proposed many-shot in-context fine-tuning (ManyICFT) with baseline methods on ICL performance. The datasets, baselines, and evaluation metrics are listed below.  

{\bf 43 Datasets:} The experimental setup follows the MetaICL experiment  \citep{Min2022metaicl} and utilizes a large collection of tasks taken
from CROSSFIT \citep{ye2021crossfit}. A total of 43 unique datasets are used, covering five task categories: text classification (CLS), question answering (QA), natural language inference (NLI), multi-label classification (ML-CLS) and multilingual summarization (SUM).  Detailed descriptions of datasets split are in Appendix Section A.1 and Table~\ref{tab:datasets}, and the in-context example preprocessing can be found in Appendix Section A.2. The evaluation metrics are in Appendix  Section A.3.

{\bf Baselines:} The baseline methods for comparison are summarized in Table~\ref{tab:baseline}. Many-shot and few-shot fine-tuning include two variants: \textit{mask all targets} and \textit{mask last target}. Additionally, the many-shot fine-tuning setting includes a standard \textit{autoregressive} variant that computes loss over all tokens. The training datasets for zero/few/many-shot are held-in datasets \footnote{The full datasets are listed in Appendix Section A}.  In contrast, task-level fine-tuning is performed on held-out datasets (14 separate models, one per task), which serve as an upper bound performance, since they are trained directly on the held-out tasks. We provide a definition for held-in and held-out data split. {\bf Held-in:} 30 training datasets for zero/few/many-shot fine-tuning, these are listed in the training column of Table~\ref{tab:datasets}. 
 {\bf Held-out:} 14 datasets for evaluation across all methods. The full datasets are listed in the test column of Table~\ref{tab:datasets}. 
For multilingual summarization task, we do fine-tuning on the high-resource language and eval on the low-resource datasets.
\begin{figure*}[ht] 
    \centering
    \subfloat[Clinc.]{\includegraphics[width=1.7in]{ 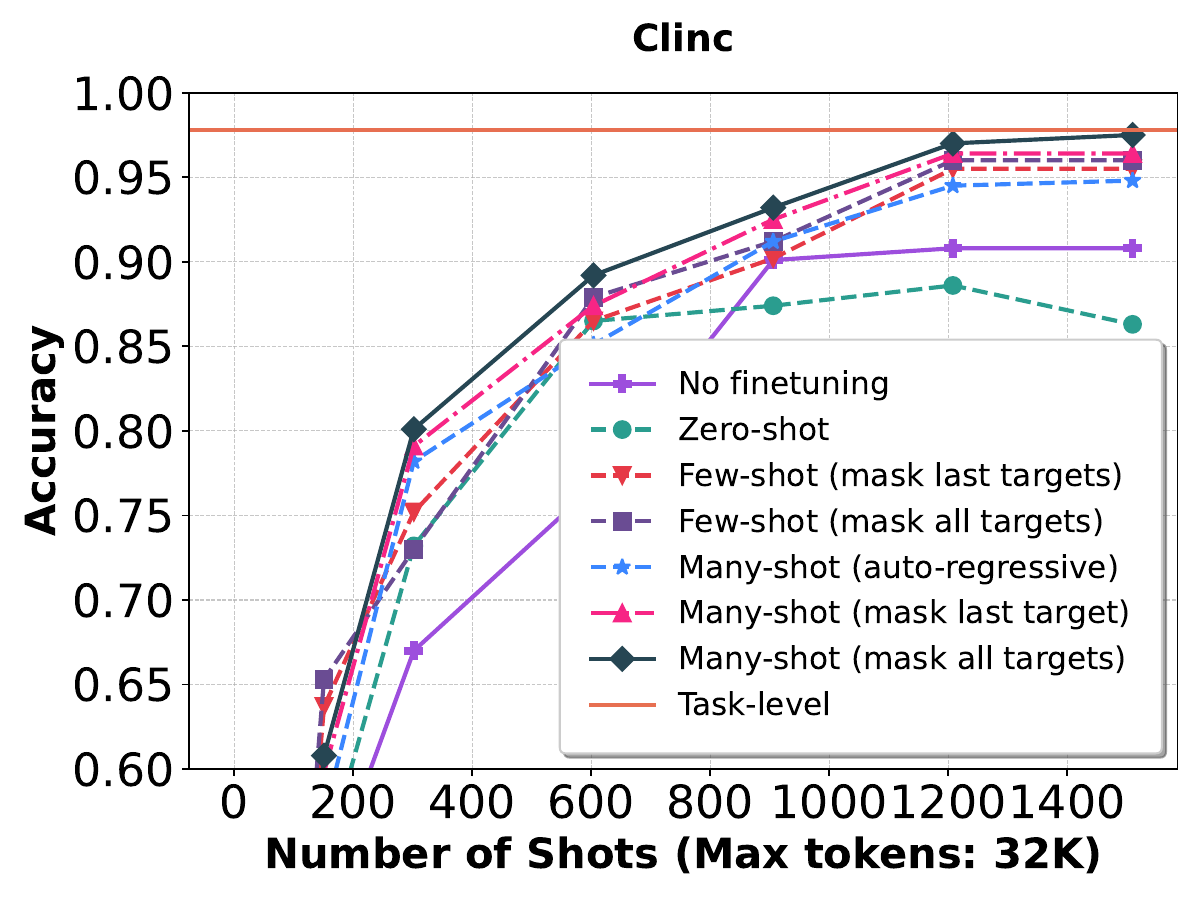}}
    \subfloat[Banking77.] {\includegraphics[width=1.7in]{ 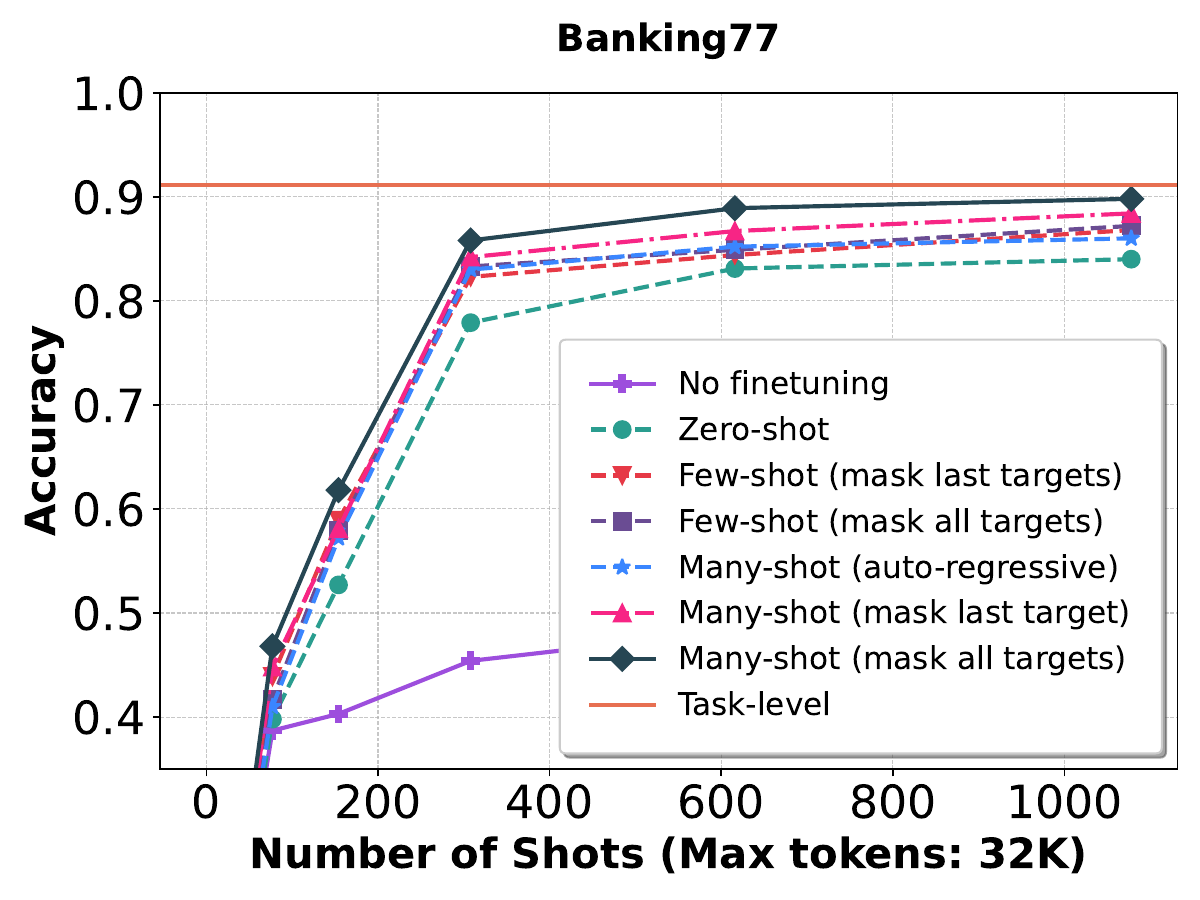}}  
    \subfloat[NLI (4 tasks in 
 Table {\ref{tab:datasets}}).]{\includegraphics[width=1.7in]{ 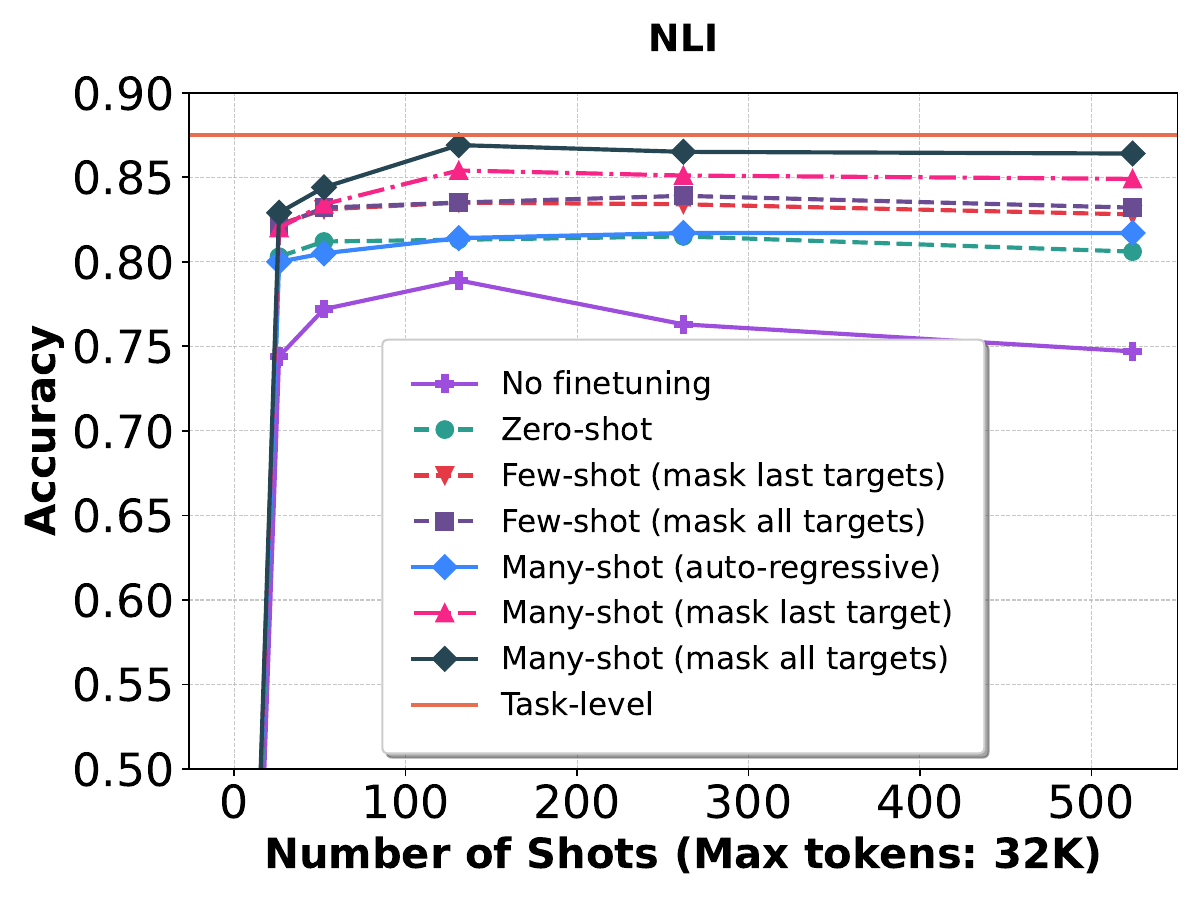}} \\
    \subfloat[QA (6 tasks in Table {\ref{tab:datasets}}).] {\includegraphics[width=1.7in]{ 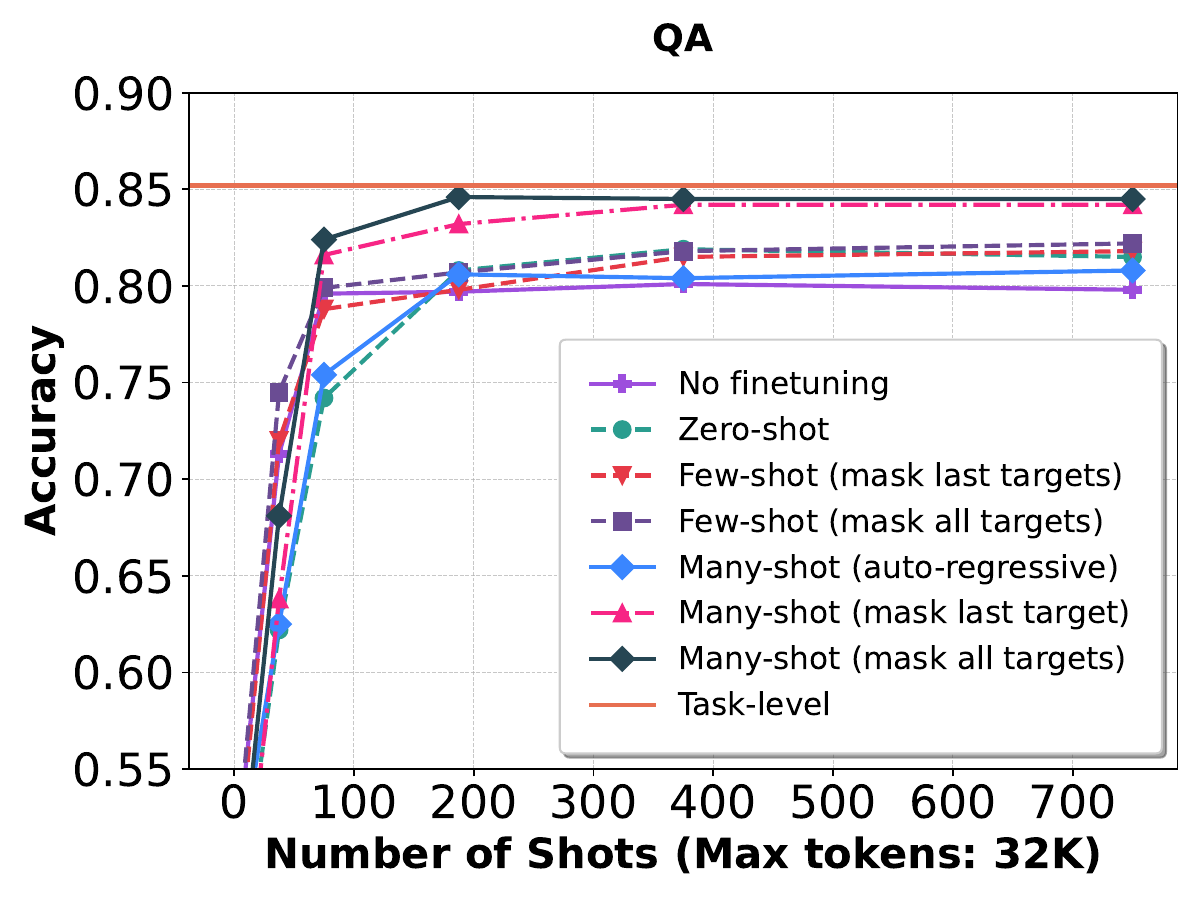}} 
        \subfloat[XLSUM PT.] {\includegraphics[width=1.7in]{ 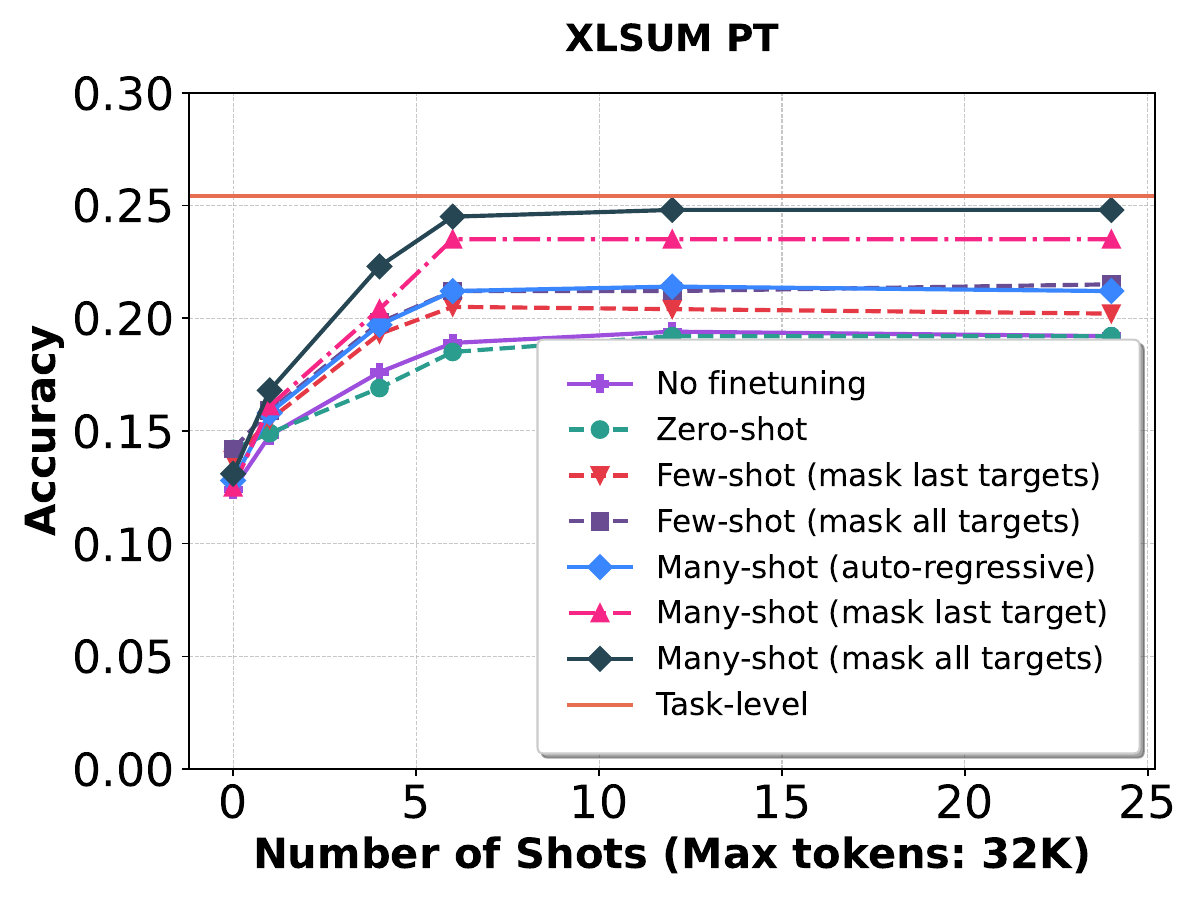}} 
    \subfloat[Multi-label classification.]{\includegraphics[width=1.7in]{ 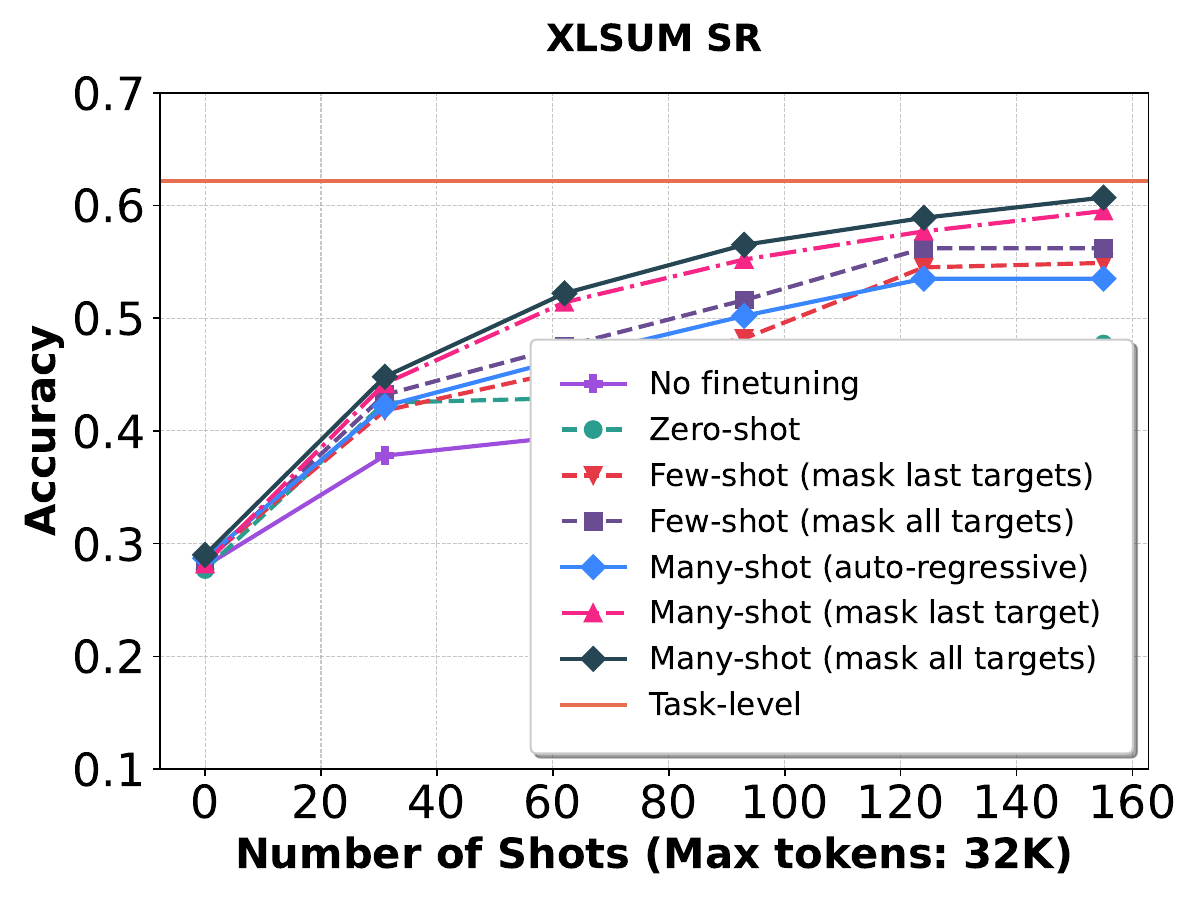}} \\
    \caption{ {\bf Scaling test on the number of shots.} Many-shot fine-tuning achieves superior performance in both few-shot and many-shot conditions. For CLS, when the number of shots is around 1.5K, many-shot fine-tuning achieves comparable performance to the task-level fine-tuning (red line). Similarly for other tasks, many-shot fine-tuning reduces the gap between ICL and task-level fine-tuning.}
    \label{fig:fivetaskres}
\end{figure*}

\begin{table*}[ht]\scriptsize
\centering
\caption{{\bf Baseline configurations.} The  tokens mask strategy, test datasets, and number of shots in fine-tuning are listed below. }

\begin{tabular}{|c|c|c|c|}
\hline
      {\bf Fine-tuning Method}                      & \begin{tabular}[c]{@{}c@{}} {\bf In-context} \\ {\bf Fine-tuning Tokens}   \end{tabular}                                                 & {\bf Training datasets}                                                                       & \begin{tabular}[c]{@{}c@{}} {\bf Number of Shots} \\ {\bf Used per Sample} \\ {\bf in Fine-tuning} \end{tabular}  \\ \hline
No fine-tuning                          & N/A                                                         & N/A                                                                        & N/A                                                                                                            \\ \hline
Task-level                 & Last                                                         & Held-out                                                           & 0                                                                                                              \\ \hline
Zero-shot                & Last                                                         &  \multirow{5}{*}{Held-in}                                       & 0                                                                                                              \\ \cline{1-2} \cline{4-4}
\multirow{2}{*}{Few-shot}  & \begin{tabular}[c]{@{}c@{}}Last\end{tabular}                                                   &                            & \multirow{2}{*}{5}                                                                                             \\ \cline{2-2}
                                      &  All   &                                                                            &                                                                                                                \\ \cline{1-2} \cline{4-4} 
\multirow{2}{*}{\textbf{Many-shot (ours)}} & Last                                                    &                                                                            & \multirow{2}{*}{ 20+} \\ \cline{2-2}
                                      & All                                                    &                                                                            &                                                                                                                \\ \hline
\end{tabular}
\label{tab:baseline}
\end{table*}

\begin{table*}\tiny
\centering
\caption{\textbf{Many-shot fine-tuning outperforms the zero/few-shot fine-tuning approaches.} Metrics are reported as both the best accuracy and the accuracy achieved with the maximum number of in-context samples (formatted in the table as Best Accuracy / Max Context Accuracy). \textbf{Bold} highlights the best performance across different meta-learning approaches (zero-, few-, and many-shot). The last row shows the performance of the dedicated fine-tuned model for each task. For NLI and QA, we report the average performance within the domain, and the concrete  tasks in evaluation are listed in Table \ref{tab:datasets}.}
\label{tab:exp}

\begin{tabular}{|c|ll|c|c|c|c|}
\hline
                              & \multicolumn{2}{c|}{\textbf{CLS}}                                & \multirow{2}{*}{\textbf{NLI}}  & \multirow{2}{*}{\textbf{QA}}  & \multicolumn{2}{c|}{\textbf{Multilingual SUM}}    \\ 
\cline{2-3}
\cline{6-7}
                              & \multicolumn{1}{l|}{\bf Clinc}               & {\bf Banking77}            &                &                 &     PT           &    SR             \\ 
\hline
\begin{tabular}[c]{@{}c@{}}No\\fine-tuning\end{tabular}                    
& \multicolumn{1}{l|}{0.908/0.908}          & 0.558/0.558          
& 0.789/0.747                               & 0.801/0.798          
& 0.195/0.193                               & 0.184/0.180          \\ 
\hline
\begin{tabular}[c]{@{}c@{}}Zero\\shot\end{tabular}                   
& \multicolumn{1}{l|}{0.886/0.863}          & 0.840/0.840          
& 0.814/0.806                               & 0.818/0.813          
& 0.192/0.192                               & 0.197/0.197           \\ 
\hline
\begin{tabular}[c]{@{}c@{}}Few-shot\\(mask last target)\end{tabular}   
& \multicolumn{1}{l|}{0.955/0.955}          & 0.868/0.868          
& 0.832/0.824                               & 0.818/0.818          
& 0.206/0.202                               & 0.214/0.214           \\ 
\hline
\begin{tabular}[c]{@{}c@{}}Few-shot\\(mask all targets)\end{tabular}   
& \multicolumn{1}{l|}{0.961/0.961}          & 0.872/0.872          
& 0.839/0.832                               & 0.822/0.822          
& 0.215/0.215                               & 0.218/0.218           \\ 
\hline
\begin{tabular}[c]{@{}c@{}}Many-shot \\ (autoregressive)\end{tabular}   
& \multicolumn{1}{l|}{0.950/0.950}          & 0.864/0. 864         
& 0.817/0.817                              & 0.808/0.808         
& 0.217/0.214                               & 0.208/0.208           \\ 
\hline
\begin{tabular}[c]{@{}c@{}}{\bf Many-shot (ours)}\\(mask last target)\end{tabular} 
& \multicolumn{1}{l|}{0.968/0.968}          & {0.884/0.884} 
& 0.855/0.849                               & 0.840/0.840          
& 0.235/0.235                               & 0.225/0.225           \\ 
\hline
\begin{tabular}[c]{@{}c@{}}{\bf Many-shot (ours)}\\(mask all targets)\end{tabular}   
& \multicolumn{1}{l|}{\textbf{0.975/0.975}} & \textbf{0.898/0.898}
& \textbf{0.868}/0.865                      & \textbf{0.845}/0.845 
& \textbf{0.248}/0.248                      & \textbf{0.239}/0.239  \\ 
\hhline{|=|=|=|=|=|=|=|}
\begin{tabular}[c]{@{}c@{}}Task-level\\fine-tuning\end{tabular}
& \multicolumn{1}{l|}{0.978}                & 0.912                
& 0.875                                     & 0.852                
& 0.253                                     & 0.245               \\ 
\hline
\end{tabular}
\end{table*}

Zero-shot fine-tuning follows the standard multi-task learning approach. In all experiments, we use Mistral-v0.3 with 32K token length \citep{jiang2023mistral} and use LoRA \citep{hu2021lora} to finetune the model. These are implemented  with PyTorch and Transformers library. The LoRA parameters are listed below. The evaluation metric and in-context sample selection are  in Appendix Section A. 

{\bf Hyperparameters:}
{\color{black}
 We tuned hyperparameters (e.g., learning rate, weight decay, LoRA rank) for each baseline on a separate validation dataset. To ensure consistency across experiments, we fixed the batch size to 1 per rank due to memory constraints, used the same optimizer (default settings), and applied a cosine learning rate decay schedule for all models.} All experiments are conducted on $8$ A100-80G GPUs, all with a per-device batch size of $1$ and a global batch size of $32$, using $4$ gradient accumulation steps. We use Adam optimizer with weight decay $10^{-5}$, $\beta_1, \beta_2$ is $0.9$ and $0.99$.

\subsection{Overall performance comparison}

We present results evaluating model accuracy from 0-shot to the maximum number of shots within the 32k context length (Mistral 7B). Table~\ref{tab:exp} highlights optimal many-shot inference, showing each model's peak accuracy (left) and accuracy at maximum context (right). Many-shot fine-tuning surpasses the base LLM and zero-/few-shot approaches on all five tasks. The \textit{mask-all targets}  strategy further boosts many-shot fine-tuning efficiency and performance. Detailed plots in subsequent sections illustrate accuracy versus the number of in-context samples.


\textbf{CLS:} Figure~\ref{fig:fivetaskres}(a-b) shows Classification (CLS) results on clinc (151 classes) and Banking77 (77 classes). While few-shot fine-tuning can outperform many-shot with less than 200 examples (consistent with MetaICL~\cite{Min2022metaicl}), many-shot consistently excels with more samples. It reaches 0.968 peak accuracy (1500 examples), outperforming few-shot methods and validating our strategy. Notably, with >1200 examples, many-shot models on held-out data near task-level fine-tuning performance

\textbf{NLI:}
Figure~\ref{fig:fivetaskres}(c) presents the experimental results for NLI task. The model is evaluated using varying numbers of in-context examples, ranging from 0 to 500 shots. Many-shot fine-tuning with the mask all targets strategy achieves the highest accuracy of 0.868 with 150 shots, outperforming the few-shot and zero-shot fine-tuning strategies by $3\%$. This improvement further narrows the performance gap between in-context learning (ICL) and task-level fine-tuning.

\textbf{QA:} Figure~\ref{fig:fivetaskres}(d) shows the experimental results for the QA task,  evaluated with  0 to 750 in-context examples. Notably, many-shot fine-tuning using the mask-all targets strategy achieves the highest performance, reaching an accuracy of 0.845 with 200 shots, which is $2.5\%$ higher than the best results obtained from few-shot and zero-shot fine-tuning. These results highlight the effectiveness of many-shot fine-tuning for QA tasks.

\textbf{Multilingual SUM:} The results for the multilingual summarization dataset are presented in Figure~\ref{fig:fivetaskres}(e) and (f).  The performance of various fine-tuning strategies are evaluated across a range of 0 to 24 shots. It is observed that many-shot fine-tuning consistently achieves the best performance across all shots. The accuracy at the maximum number of shots is $0.467$, which is $1\%$ higher than the  results of the few-shot fine-tuning method and $3.7\%$ higher than no-finetuned model.



\subsection{Long-context capability comparison}

\begin{table*}\tiny
\centering
\caption{\color{black}{\bf  Long-context capability comparison for different fine-tuning methods.} Fine-tuning method comparison for long-context generation. Zero- and few-shot approaches exhibit catastrophic forgetting, which many-shot fine-tuning alleviates through explicit long-context meta-training}
\label{tab:pg19}
\begin{tabular}{|c|c|c|c|c|c|c|c|c|c|}
\hline
\begin{tabular}[c]{@{}c@{}}Context\\length\end{tabular}    & 32K                                                     & 16K                                                       & 8K                                                       & 4K                                                       & 1K                                                       & 512                                                       & 256                                                        & 128                                                        & 32                                                        \\ \hline
\begin{tabular}[c]{@{}c@{}}No\\Fine-tuning\end{tabular}     & 7.38                                                    & 7.55                                                      & 7.78                                                     & 8.07                                                     & 9.12                                                     & 10.10                                                     & 11.81                                                      & 12.95                                                      & 14.48                                                     \\ \hline
\begin{tabular}[c]{@{}c@{}}Zero\\shot\end{tabular}      & \begin{tabular}[c]{@{}c@{}}7.49\\ (+ 0.11)\end{tabular} & \begin{tabular}[c]{@{}c@{}}7.77	\\ (+ 0.22)\end{tabular}  & \begin{tabular}[c]{@{}c@{}}8.05	\\ (+ 0.27)\end{tabular} & \begin{tabular}[c]{@{}c@{}}8.25 \\ (+ 0.18)\end{tabular} & \begin{tabular}[c]{@{}c@{}}9.39 \\ (+ 0.27)\end{tabular} & \begin{tabular}[c]{@{}c@{}}10.32 \\ (+ 0.22)\end{tabular} & \begin{tabular}[c]{@{}c@{}}11.96  \\ (+ 0.15)\end{tabular} & \begin{tabular}[c]{@{}c@{}}13.07  \\ (+ 0.12)\end{tabular} & \begin{tabular}[c]{@{}c@{}}14.60 \\ (+ 0.12)\end{tabular} \\ \hline
\begin{tabular}[c]{@{}c@{}}Few-shot\\(mask all targets)\end{tabular}     & \begin{tabular}[c]{@{}c@{}}7.51\\ (+ 0.13)\end{tabular} & \begin{tabular}[c]{@{}c@{}}7.73	\\ (+ 0.18)\end{tabular}  & \begin{tabular}[c]{@{}c@{}}8.01	\\ (+ 0.23)\end{tabular} & \begin{tabular}[c]{@{}c@{}}8.27 \\ (+ 0.20)\end{tabular} & \begin{tabular}[c]{@{}c@{}}9.35 \\ (+ 0.23)\end{tabular} & \begin{tabular}[c]{@{}c@{}}10.27 \\ (+ 0.17)\end{tabular} & \begin{tabular}[c]{@{}c@{}}12.04 \\ (+ 0.23)\end{tabular}  & \begin{tabular}[c]{@{}c@{}}13.07 \\ (+ 0.12)\end{tabular}  & \begin{tabular}[c]{@{}c@{}}14.62 \\ (+ 0.14)\end{tabular} \\ \hline
\begin{tabular}[c]{@{}c@{}}{\bf Many-shot (Ours)}\\(mask all targets)\end{tabular}    & \begin{tabular}[c]{@{}c@{}}7.45\\ (+ 0.07)\end{tabular} & \begin{tabular}[c]{@{}c@{}}7.61	\\ (+0.06)\end{tabular}   & \begin{tabular}[c]{@{}c@{}}7.88	\\ (+ 0.10)\end{tabular} & \begin{tabular}[c]{@{}c@{}}8.16 \\ (+ 0.11)\end{tabular} & \begin{tabular}[c]{@{}c@{}}9.26 \\ (+ 0.14)\end{tabular} & \begin{tabular}[c]{@{}c@{}}10.19 \\ (+ 0.09)\end{tabular} & \begin{tabular}[c]{@{}c@{}}11.93 \\ (+ 0.12)\end{tabular}  & \begin{tabular}[c]{@{}c@{}}13.04 \\ (+0.09)\end{tabular}   & \begin{tabular}[c]{@{}c@{}}14.58 \\ (+ 0.07)\end{tabular} \\  
\hhline{|=|=|=|=|=|=|=|=|=|=|}
\begin{tabular}[c]{@{}c@{}}Many-shot \\(auto-regressive)\end{tabular}  & \begin{tabular}[c]{@{}c@{}}7.42\\ (+ 0.04)\end{tabular} & \begin{tabular}[c]{@{}c@{}}7.60	 \\ (+ 0.05)\end{tabular} & \begin{tabular}[c]{@{}c@{}}7.83	\\ (+ 0.05)\end{tabular} & \begin{tabular}[c]{@{}c@{}}8.14 \\ (+ 0.07)\end{tabular} & \begin{tabular}[c]{@{}c@{}}9.20 \\ (+ 0.08)\end{tabular} & \begin{tabular}[c]{@{}c@{}}10.18 \\ (+ 0.08)\end{tabular} & \begin{tabular}[c]{@{}c@{}}11.83 \\ (+ 0.02)\end{tabular}  & \begin{tabular}[c]{@{}c@{}}13.02\\ (+ 0.07)\end{tabular}   & \begin{tabular}[c]{@{}c@{}}14.51 \\ (+ 0.03)\end{tabular} \\ \hline
\end{tabular}
\end{table*}

We investigated how different fine-tuning strategies affect LLMs' general capabilities, focusing on long-context understanding. LLMs are known to suffer from catastrophic forgetting during fine-tuning \citep{zhai2023investigating, luo2023empirical}. To evaluate this part, we use the PG-19 validation set \citep{rae2019compressive} and report the perplexity of each method on this dataset, as shown in Table~\ref{tab:pg19}. Perplexity score at various context length has been a standard evaluation for long-context capability evaluation.

The results indicate that autoregressive fine-tuning with many-shot training datasets leads to increased perplexity, due to distributional mismatch between the PG-19 dataset and the fine-tuning corpus. More notably, zero-shot and few-shot fine-tuning result in significant degradation in performance, with perplexity increasing by 0.3 at 4K context length and by 0.2 at 32K context length compared to the non-fine-tuned baseline. This confirms that both zero-shot and few-shot fine-tuning can lead to noticeable catastrophic forgetting. In contrast, our many-shot in-context fine-tuning approach effectively mitigates this issue through meta-training.

\subsection{Ablation study}

\begin{figure*}[h]
    \centering
    \subfloat[Ablation study on CLS.]{\includegraphics[width=1.7in]{ 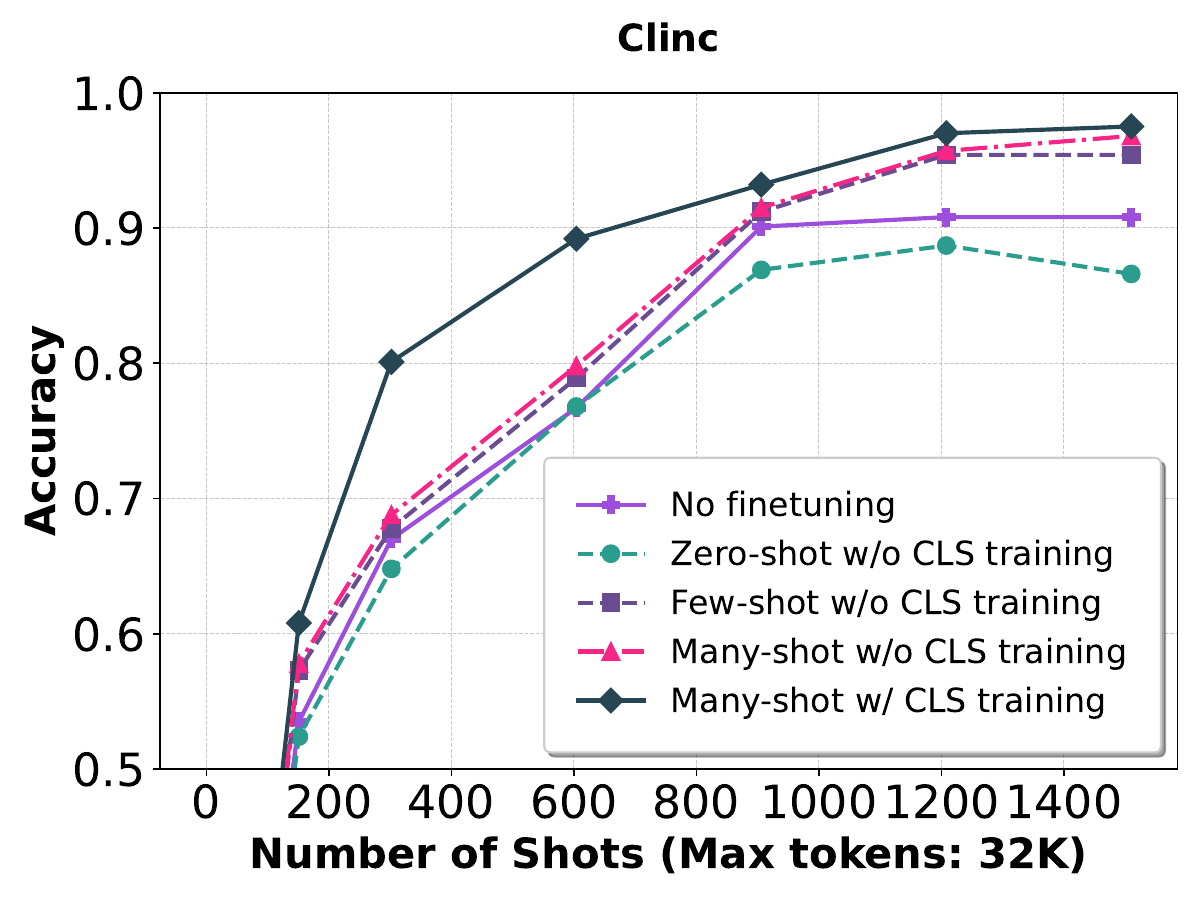}} 
    \subfloat[Ablation study on CLS.]{\includegraphics[width=1.7in]{ 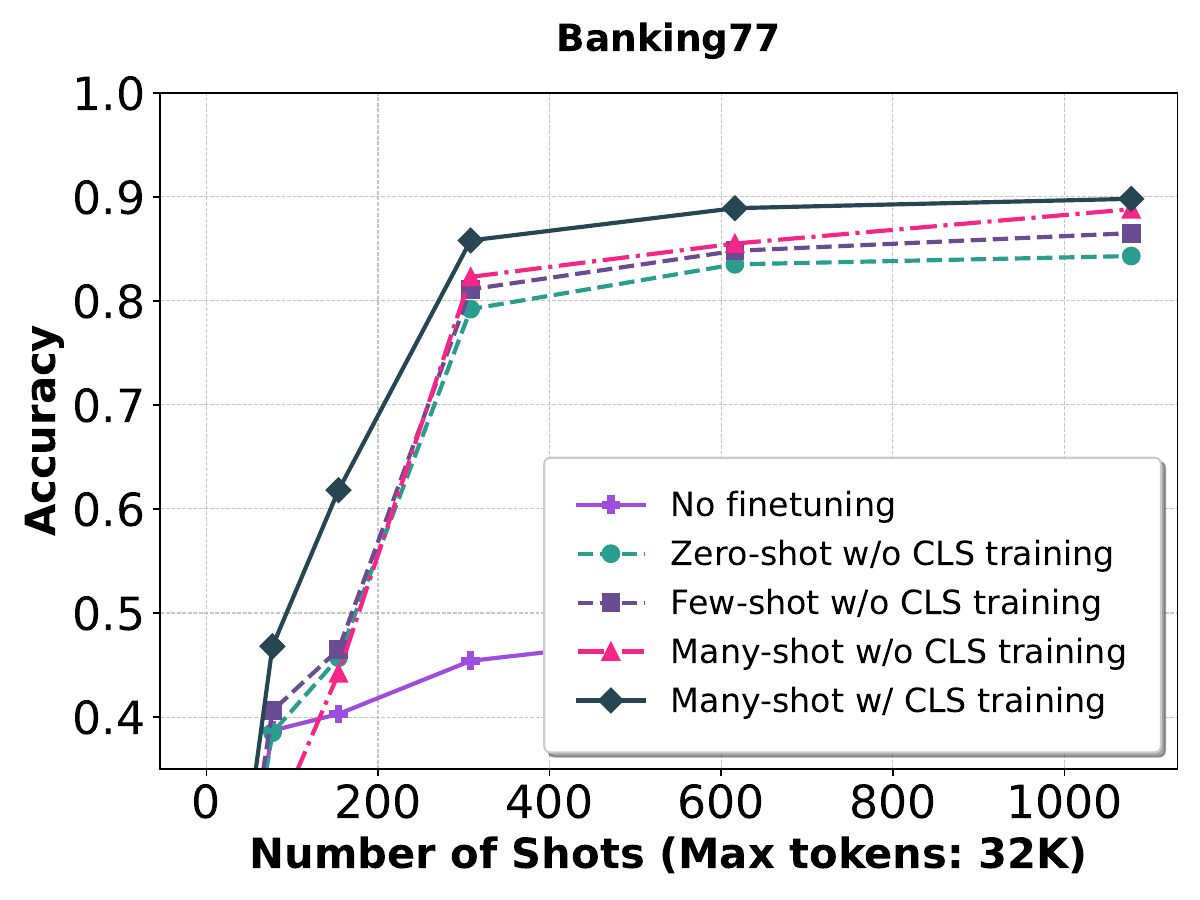}} 
    \subfloat[Ablation study on NLI.]{\includegraphics[width=1.7in]{ 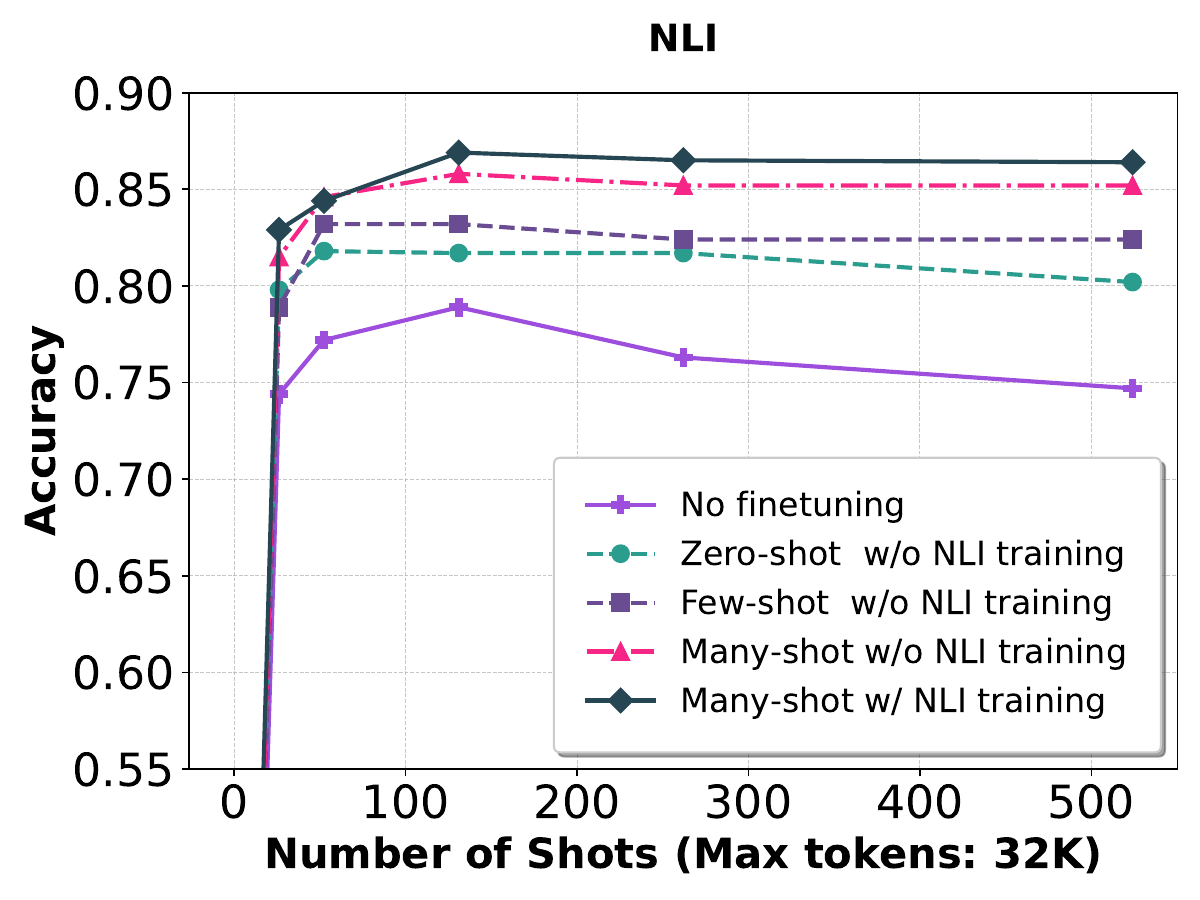}} \\
    \subfloat[Ablation study on QA.]{\includegraphics[width=1.7in]{ 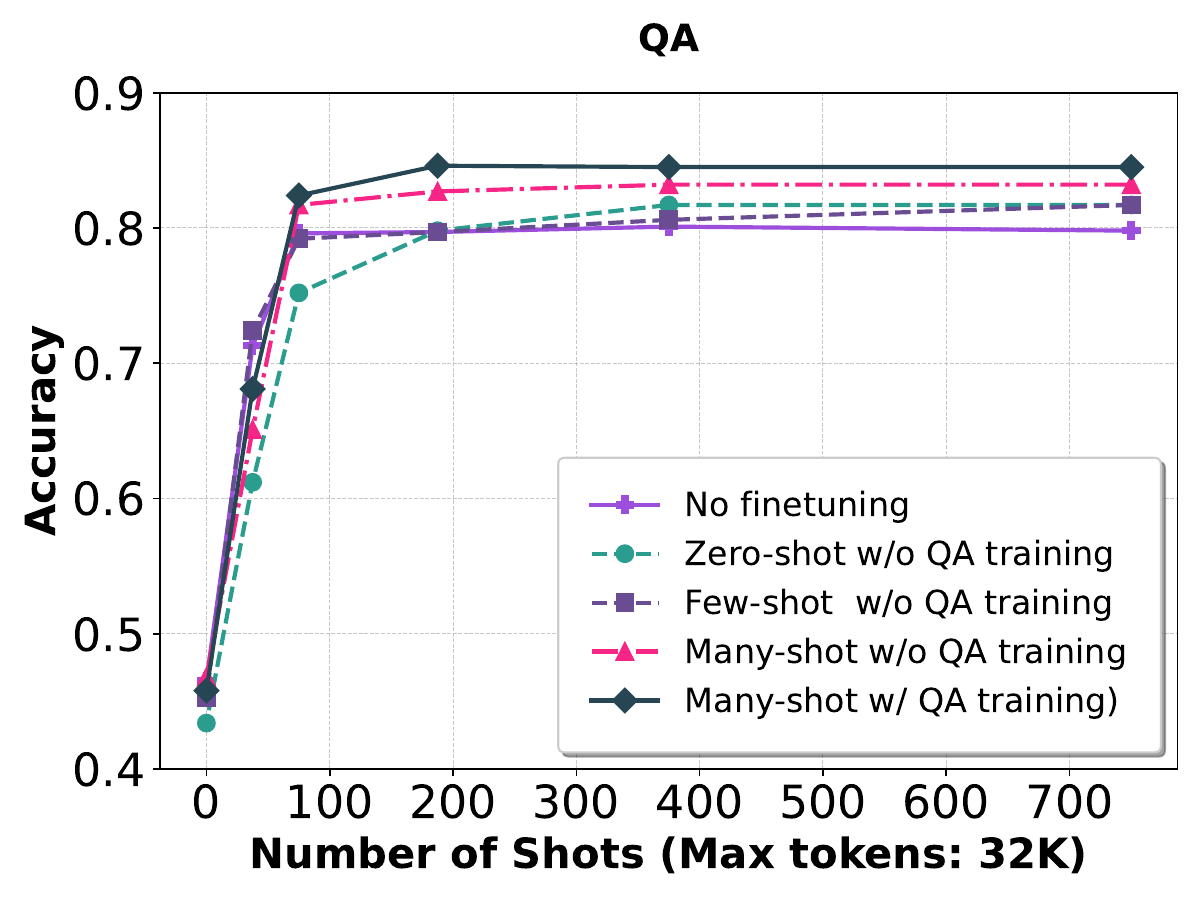}} 
    \subfloat[Ablation study on XLSUM PT.]{\includegraphics[width=1.7in]{ 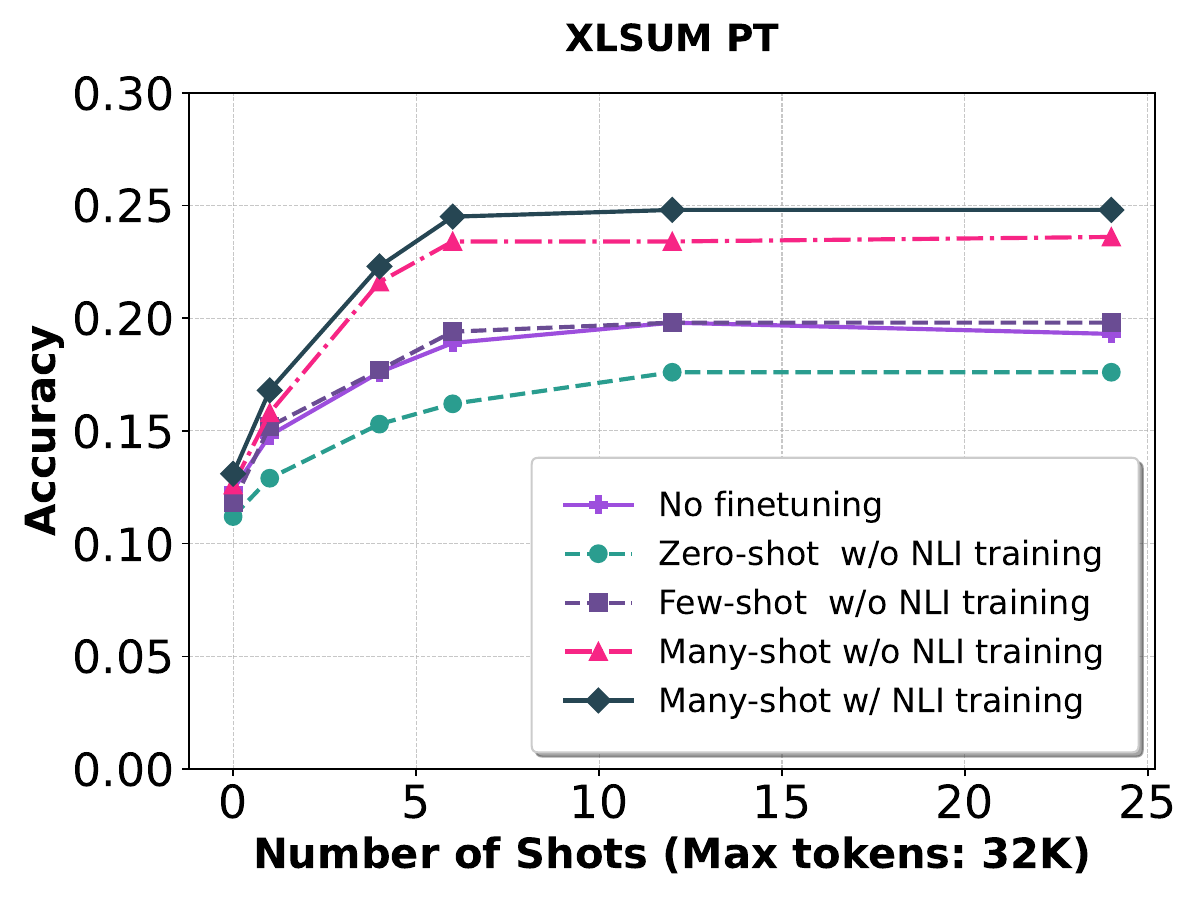}} 
    \subfloat[Ablation study on XLSUM SR.]{\includegraphics[width=1.7in]{ 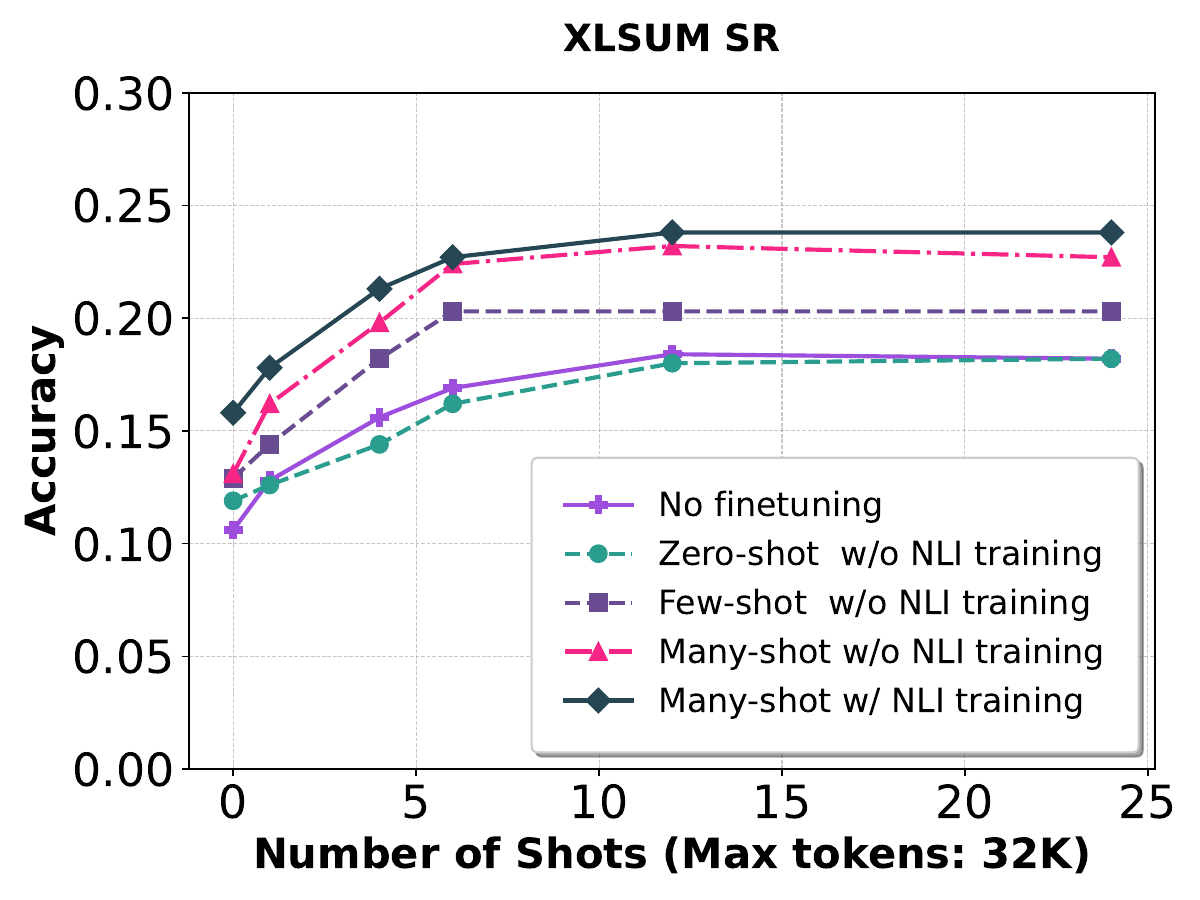}} 
    \caption{\color{black} {\bf Ablation study} is done by removing the corresponding training dataset of the test task. }
    \label{fig:abalation}
\end{figure*}

The goal of this ablation study is to evaluate the generalization ability of many-shot fine-tuning across diverse tasks. We systematically evaluate the performance of our fine-tuned model on five held-out test tasks. For each target task, we exclude all training datasets belonging to the same task category during fine-tuning. This setup enables us to examine whether the model can perform many-shot in-context learning on unseen tasks, without exposure to in-domain training tasks. The results are summarized in Figure~\ref{fig:abalation} and Table~\ref{tab:abalationtab} (Appendix Section C.1).

Specifically, for the {\bf CLS} task (evaluated on the Banking77 dataset), we remove all classification-related training datasets. As shown in Figures~\ref{fig:abalation}(a) and ~\ref{fig:abalation}(b), the fine-tuned model underperforms the base model when the number of shots is fewer than 200, but significantly outperforms it at larger shot sizes. This suggests that many-shot ICL capabilities learned from non-CLS tasks can transfer to classification tasks when sufficient context is provided. In addition, comparing many-shot and zero/few-shot fine-tuning (accuracy is reported in Table 9 of Appendix section C.1), we observe that zero-shot fine-tuning performs worse than the base model without fine-tuning, indicating that removing CLS-related training data induces catastrophic forgetting.

For the {\bf NLI} task (Figure~\ref{fig:abalation}(c)), excluding NLI training datasets has little impact on performance. The ablation model’s curve (black line) nearly overlaps with the original model’s, demonstrating that many-shot ICL capabilities learned from other tasks generalize well to NLI. In the {\bf QA} setting (Figure~\ref{fig:abalation} (d)), the model fine-tuned without QA data still consistently outperforms the base model, suggesting strong transfer. However, incorporating QA data during fine-tuning can yield additional performance gains. 
For the {\bf Multilingual SUM} task in Figure~\ref{fig:abalation}(e) and (f), a similar pattern emerges: many-shot fine-tuning without direct MATH supervision still improves performance over the base model. We attribute this to transfer from other CoT-style (Chain-of-Thought) reasoning tasks in the training set.

These findings suggest that many-shot fine-tuning enables robust transfer to CLS, QA, Multilingual SUM, and NLI, even without task-specific training data. 


\section{Conclusion}
In this paper, we introduce ManyICFT, a novel meta-training framework that enhances many-shot ICL for LLMs. Our approach shows superior performance in both few-shot and many-shot scenarios across diverse downstream tasks. Extensive experiments demonstrate that ManyICFT outperforms conventional zero-shot and few-shot fine-tuning methods while mitigating catastrophic forgetting. These findings highlight ManyICFT’s potential to streamline LLM deployment by reducing reliance on task-specific fine-tuning.

However, challenges remain. Limited context windows and the token demands of many-shot prompts can constrain space for system and user inputs and slow inference. Future work will explore optimizing context length, reducing inference costs, and extending ICL capabilities to more domains.

\bibliography{main}
\appendix
\include{appendix}
\end{document}

%% file: appendix.tex
\section{Datasets}
\subsection{Datasets split}
The training and test datasets are listed in Table~\ref{tab:datasets}. We used 43 datasets in total, including 29 training datasets and 14 test datasets among $5$ tasks. There is no overlap between the training and test datasets. For the training datasets, we select the training samples from the training split, and for the test datasets, we select the test samples from the test split of each dataset.

For many-shot fine-tuning, a total of 20K training instances are used (4K per task), with each instance composed of the maximum number of in-context examples that can fit within the model’s input context length. For few-shot fine-tuning setting, 5-shot prompts are applied across all datasets. On average, each training instance contains 212 in-context examples. This results in approximately 40 times more sequences for few-shot fine-tuning with the mask-all-targets strategy and 200 times more for zero-shot and few/many-shot fine-tuning with the mask-last-target strategy. This sampling approach ensures all baseline methods receive an equivalent number of training samples, enabling fair comparison across zero-shot, few-shot, and many-shot fine-tuning. For each instance, in-context examples are randomly selected from the training split. Datasets for each task are partitioned into non-overlapping training and test sets, allowing for evaluation of held-out generalization performance across all methods

\subsection{ICL examples}

{$\bullet$ In-context sample selection:} For the classification task, we use stratified sampling from training data, which samples the same number of samples for each class to ensure the balance of classes. For other tasks, we do random sampling from the training data. In-context samples are not fixed for different samples.

{\color{black}
{$\bullet$ Prompt construction: }
We construct the  ICL prompts by concatenating the maximum number of input-output pairs that can fit within the model’s context window. The following illustrates the prompt template used for the Clinc dataset with a 3-shot example. Prompts for other datasets follow a similar construction methodology.

\begin{quote}
\begin{verbatim}
Task: You will receive a text input. Your task is to classify the 
intent of the input text. Here are some examples:

Text: Let me know how much money I will need to spend on paying bills
Intent: Bill balance

Text: Could you let me know what song this is
Intent: What song

Text: Should I take a different route to work
Intent: Traffic

Text: {query}
Intent:
\end{verbatim}
\end{quote}
}

\subsection{Evaluation metrics}

For classification, QA, NLI and paraphrase detection tasks, we report the  accuracy, and for summarization task, we report the ROUGE-L between prediction and ground-truth, which measures the longest overlap ratio between two texts.

\begin{table*}[]\scriptsize
\centering
\caption{{\bf Datasets configurations.} Training and test datasets for experiments. }
\begin{tabular}{|l|l|l|}
\hline
                                                             & \begin{tabular}[c]{@{}l@{}}{\bf Training} {\bf (29 datasets)}\end{tabular}                                                                       & \begin{tabular}[c]{@{}l@{}}{\bf Test}  {\bf (14 datasets)}\end{tabular}                                                                    \\ \hline
\begin{tabular}[c]{@{}l@{}}CLS \\ Training: 10 datasets \\ Test: 2 datasets \end{tabular}  & \begin{tabular}[c]{@{}l@{}}hate\_speech\_offensive \citep{de2018hate} \\ google\_wellformed\_query \citep{faruqui2018identifying} \\ tweet\_eval-emoji \citep{barbieri2020tweeteval} \\ ade\_corpus\_v2-classification \citep{gurulingappa2012development} \\ glue\_sst \citep{wang2018glue}\\ ag\_news \citep{gulli2005corpus} \\ paws \citep{zhang2019paws}\\ glue\_qqp \citep{wang2018glue} \\ 
bbh\_geometry\_shapes \citep{suzgun2022challenging} \\
bbh\_logic\_deduction\_five\_objects \citep{suzgun2022challenging}
\\\end{tabular} & \begin{tabular}[c]{@{}l@{}}clinc \\ \citep{rabinovich2022reliable}\\ banking77 \\ \citep{casanueva2020efficient}\end{tabular}                                                                       \\ \hline
\begin{tabular}[c]{@{}l@{}}NLI \\ Training: 6 datasets \\ Test: 4 datasets  \end{tabular} & \begin{tabular}[c]{@{}l@{}}circa \citep{louis2020d}\\ scitail \citep{khot2018scitail}\\ art \citep{bhagavatula2019abductive}\\ glue\_qnli \citep{rajpurkar2016squad}\\ glue\_mnli \citep{williams2017broad}\\ anli \citep{nie2019adversarial}\end{tabular}                                                                          & \begin{tabular}[c]{@{}l@{}}glue\_rte \\ \citep{dagan2005pascal}\\ glue\_wnli \\ \citep{levesque2012winograd}\\ sick \citep{marelli2014semeval}\\ super\_cb \\ \citep{dolan2005automatically}\end{tabular}                                                \\ \hline
\begin{tabular}[c]{@{}l@{}}QA \\ Training: 8 datasets \\ Test: 6 datasets \end{tabular}  & \begin{tabular}[c]{@{}l@{}}hellaswag \citep{zellers2019hellaswag}\\ cosmos\_qa \citep{huang2019cosmos}\\ kilt\_ay2 \citep{hoffart2011robust}\\ quail \citep{rogers2020getting}\\ dream \citep{sun2019dream}\\ sciq \citep{welbl2017crowdsourcing}\\ qasc \citep{khot2020qasc} \\ gpqa \citep{rein2023gpqa}\end{tabular}                                                                & \begin{tabular}[c]{@{}l@{}}quartz\_no\_knowledge \\ \citep{tafjord2019quartz}\\ quartz\_with\_knowledge \\ \citep{tafjord2019quartz}\\ openbookqa \citep{mihaylov2018can}\\ ai2\_arc \citep{clark2018think}\\ codah \citep{chen2019codah}\\ quarel \citep{tafjord2019quarel}\end{tabular} \\ \hline

\begin{tabular}[c]{@{}l@{}}Multilingual SUM \\ Training: 7 datasets \\ Test: 2 datasets \end{tabular} & \begin{tabular}[c]{@{}l@{}} xlsum en, xlsum zh, xlsum jp, \\ xlsum fr,  xlsum ru, xlsum es, xlsum ko \citep{hasan-etal-2021-xl} \end{tabular}                                                                                     & \begin{tabular}[c]{@{}l@{}} xlsum pt \citep{hasan-etal-2021-xl} \\ xlsum sr \citep{hasan-etal-2021-xl}
\end{tabular}                                                                                                                          \\ \hline
\end{tabular}
\label{tab:datasets}
\end{table*}

\section{Computational Cost Analysis}


\subsection{Training cost}

We compare mask all targets training cost with that of mask last token in terms of complexity analysis and quantitative estimation. 
\subsubsection{Complexity analysis}
 Our proposed mask all targets strategy is significantly more efficient than the standard mask last target approach for training ICL capabilities. \textit{Mask last target} needs separate sequences for 1-shot, 2-shot,...n-shot, processing, with $O(n\times n_w)$ tokens per instance,  where $n$ is the maximum number of shots and $n_w$ is the maximum token length of the context window. In contrast,\textit{mask all targets} learns from 0 to n-shot from a single packed sequence, processing only $O(n_w)$ tokens. 
      
\subsubsection{Quantitative estimation} 

For the training, we use $N = 70K , n_w= 32k , n= 100$, and there are approximately 212 avg shots. In this setup, \textit{mask all targets} processed around 2.2B tokens versus 237B for \textit{mask last target}, which results in roughly a 100x reduction.
\begin{table*}[]\scriptsize
\centering
\caption{{\bf Mask all targets and mask last target training efficiency comparison.}}
\begin{tabular}{|c|c|c|}
\hline
  Training Strategy        &    Theoretical analysis            & Practical estimation  \\ \hline
Mask last target & $O(N\times n_w\times n)$ & 70K$\times$32K$\times$106$\sim$237B \\ \hline
Mask all targets  & $O(N\times n_w)$   & 70K$\times$32K $\sim$2.2B    \\ \hline
\end{tabular}
\label{tab:traincompute}
\end{table*}
We compare the computational cost of meta-training paradigm  with  mask last token in terms of complexity analysis and quantitative estimation.   ManyICFT enables a "fine-tune only once" model. While the single fine-tuning uses long sequences (handled efficiently by mask-all-targets), it avoids the significant cumulative cost of separate fine-tuning for numerous downstream tasks.  The number of training tokens are summarized in Table~\ref{tab:metatraining}.

\begin{table*}[]\scriptsize
\centering
\caption{{\bf Comparison on the number of training tokens for Meta training with task-level fine-tuning.}}

\begin{tabular}{|c|c|}
\hline
Meta learning (ManyICFT)    & Task-level fine-tuning \\ \hline
70K $\times$ 32K = 2.2B         & 4K $\times$ 8K $\times$ 1K = 32B \\ \hline
\end{tabular}
\label{tab:metatraining}
\end{table*}

{ \subsection{Inference cost}}

\begin{table*}[]\scriptsize
\centering
\caption{{\bf Inference cost comparison: the few-shot, many-shot without KV cache and with KV cache inference theoretical complexity and its cost in practice. ($k$ is the number of shots for few-shot setting and $n$ is that for many-shot setting, $n_x$ denotes the average number of input tokens and $n_y$ is that of output tokens for each shot.)}}

\begin{tabular}{|c|c|c|}
\hline
                 & Theoretical complexity                                      & \begin{tabular}[c]{@{}l@{}}Practical costs for \\ 32K context length \end{tabular} 
                   \\ \hline
Few-shot         & $O( k^2 (n_x + n_y)^2$)   & $0.0025\times$  \\ \hline
Many-shot w/o KV cache & $O( n^2 (n_x + n_y)^2 )$ & $1\times$                  \\ \hline
Many-shot w KV cache   & $O(n(n_x + n_y)^2$ )                    & $0.01\times$               \\ \hline
\end{tabular}
\label{tab:infercompute}
\end{table*}
\subsubsection{Complexity analysis}
The vanilla inference costs is quadratic to the number of context length due to attention mechanism. With the help of KV-cache, the context can be cached and reduce the inference costs linear to the context length. 
\subsubsection{Practical costs}
Assuming the baseline computational latency normalized to 1$\times$ (without KV cache), the work \cite{xiao2025efficient} (Table 2) reports that KV caching reduces latency to 0.51 times the baseline for the LLaMA2 model and 0.11 times the baseline for LLaMA3 when using a 30K context length. For theoretical comparison, we summarize the computational complexity in Table~\ref{tab:infercompute}.   Loading the KV cache from SSD to RAM incurs a one-time cost equivalent to 0.1 times the baseline latency, while the marginal per-example inference cost is reduced to 0.01 times. In practical scenarios, accounting for the system prompt, and other overheads, the effective per-example cost can be further reduced to approximately 0.0025 times the baseline. 

In conclusion,  our mask all targets strategy makes the single ManyICFT fine-tuning phase highly efficient (~$100\times$ fewer tokens processed in  our
implementation than mast last target strategy). This achieves efficient fine-tuning  than fine-tuning each individual task in task-level fine-tuning approaches.

\section{Experiment Results}
\subsection{Additional results on ablation study}
The  performance comparison among ManyICFT and baselines on many-shot inference for ablation study is reported in Table~\ref{tab:abalationtab}. This demonstrates the proposed ManyICFT still keeps best performance on ablation study. We also report zero-shot and five-shot results of ablation study in Table~\ref{tab:abalationtab0shot}. These results show that zero-shot, few-shot, and autoregressive fine-tuning perform worse than the base model, indicating significant forgetting on out-of-domain tasks. In contrast, the many-shot fine-tuned model maintains strong performance when tested on unseen task types.

\begin{table*}\scriptsize
\centering
\caption{{\bf Ablation study.} Metrics are reported in both the best accuracy/accuracy with maximum in-context samples. Ablation are done by removing the corresponding training dataset of the test task. For example, when testing the performance on Clinc and Banking77, we remove the CLS dataset during training.   } 
\label{tab:abalationtab}
\begin{tabular}{|c|ll|c|c|c|c|}
\hline
                              & \multicolumn{2}{c|}{\textbf{CLS}}                                & \multirow{2}{*}{\textbf{NLI}}  & \multirow{2}{*}{\textbf{QA}}  & \multirow{2}{*}{\textbf{MATH}}  & \multirow{2}{*}{\textbf{SUM}}  \\ \cline{2-3}
                              & \multicolumn{1}{l|}{\bf Clinc}               & {\bf BANKING77}            &                &                 &                &                 \\ \hline
\begin{tabular}[c]{@{}c@{}}No \\  fine-tuning \end{tabular}                    & \multicolumn{1}{l|}{0.908/0.908}          & 0.558/0.558          & 0.789/0.747              & 0.801/0.798           & 0.435/0.430            &  {\bf 0.301}/0.291         \\ \hline
\begin{tabular}[c]{@{}c@{}}Zero-shot \\  FT \end{tabular}                    & \multicolumn{1}{l|}{0.876/0.858}          & 0.835/0.835          & 0.808/0.798              & 0.804/0.804             & 0.435/0.435            &  0.272/0.280         \\ \hline
\begin{tabular}[c]{@{}c@{}}Few-shot \\  FT \end{tabular}                    & \multicolumn{1}{l|}{0.955/0.955 }          & 0.860/0.860          & 0.818/0.812              & 0.808/0.808        & 0.446/0.442            &  0.291/0.291         \\ \hline
    \begin{tabular}[c]{@{}c@{}} { {\bf  Many-shot } } \\ {\bf  {(Ours)}}\end{tabular}   & \multicolumn{1}{l|}{{\bf 0.964/0.964}} & {\bf 0.880/0.880} & {\bf 0.844}/{\bf 0.844}     & {\bf 0.818}/{\bf 0.814}    & {\bf 0.454}/{\bf 0.448}      & {0.296}/{\bf 0.296}     \\   
    \hhline{|=|=|=|=|=|=|=|}
    \begin{tabular}[c]{@{}c@{}} { Many-shot }  \\ (With all training data)\end{tabular}   & \multicolumn{1}{l|}{{0.968/0.968}} & {0.885/0.885} & {0.859}/0.854     & {0.835}/0.835    & {0.478}/0.467     &{0.308}/0.308       \\ \hline
\end{tabular}
\end{table*}

\begin{table*}[]\scriptsize
\label{tab:oodzerofew}
\centering
\caption{{\bf Out-of-domain generalization results for zero-shot and five-shot evaluations.} Accuracy is reported in the format of \textit{zero-shot/five-shot}. This experiment assesses the generalization ability of different fine-tuning strategies across six distinct tasks without including task-specific training data} 
\label{tab:abalationtab0shot}
\begin{tabular}{|c|c|c|c|c|c|c|}
\hline
                                      & Clinc & Banking & NLI   & QA    & GSM8K & SUM   \\ \hline
No fine-tuning                        & 0/0.535     & 0/0.387      & 0.446/0.744 & {\bf 0.465}/{\bf 0.653} &0/{\bf 0.411}      & 0.170/0.264 \\ \hline
Zero-shot                  & 0/0.514     & 0/0.372       & 0.397/0.792 & 0.424/0.577 & 0/0.385     & 0.147/0.245 \\ \hline
\begin{tabular}[c]{@{}c@{}} { Few-shot }  \\ \end{tabular}             & 0/0.563     & 0/{\bf 0.392}       & 0.415/0.779 & 0.445/0.647 & 0/0.392     & 0.152/0.271 \\ \hline
\begin{tabular}[c]{@{}c@{}} { Many-shot }  \\ (autoregressive)\end{tabular}  & 0/0.538     & 0/0.382      & 0.422/0.727 &  0.453/0.551 & 0/0.372     & 0.158/0.232 \\ \hline
\begin{tabular}[c]{@{}c@{}} { {\bf Many-shot} }  \\ (Ours)\end{tabular}                        & 0/{\bf 0.558}     & 0/0.388       & {\bf 0.498}/{\bf 0.805} & 0.451/0.632 & 0/0.404     & {\bf 0.175}/{\bf 0.274} \\ \hline
\end{tabular}
\end{table*}

\subsection{Analysis on the number of training samples}

{\color{black}
We provide a quantitative analysis for the number of training samples for baseline and proposed method. The results are shown in Table~\ref{tab:num_sample_exp} and we can observe that ManyICFT only requires  1K in-context examples for CLS tasks while standard task-level fine-tuning requires 4K fine-tuning samples to achieve the best performance. Similarly for other tasks including QA, NLI, GSM8K and SUM, ManyICFT reduces the required samples significantly. ManyICFT leverages relatively few inference-time examples (hundreds to thousands) to achieve strong results, replacing the need for potentially much larger task-specific training sets used in standard fine-tuning. While ManyICFT requires an initial diverse dataset for its one-time fine-tuning, this cost is amortized across all tasks, avoiding the cumulative effort of sourcing and training on potentially thousands of samples for each new task individually.
 }

\begin{table*}[]\scriptsize
\centering
\caption{{ \bf Comparison of the number of examples required to achieve the best performance.}}
\label{tab:num_sample_exp}
\begin{tabular}{|l|l|l|l|l|l|l|}
\hline
                      & CLS  & Banking77 & NLI   & QA   & GSM8K & SUM   \\ \hline
Task-level FT & 4K   & 4K        & 12K   & 18K  & 7.4K  & 4K    \\ \hline
Many-shot FT            & {\bf 1.2K} &{\bf 0.6K}      &{\bf  0.13K} &{\bf  0.2K} & {\bf 0.08K} &{\bf  0.02K} \\ \hline
\begin{tabular}[c]{@{}c@{}} { Relative }  \\ improvements\end{tabular}          & 3.3$\times$ & 6.6$\times$      & 92$\times$   & 90$\times$  & 92$\times$   & 200$\times$  \\ \hline
\end{tabular}
\label{tab:numofexamples}
\end{table*}




